\documentclass[a4paper, 11pt]{article}
\usepackage[sort&compress,square,comma,authoryear]{natbib}

\usepackage{a4wide}
\usepackage{algorithm}
\usepackage{algorithmic}
\usepackage[algo2e,linesnumbered,ruled]{algorithm2e}

\usepackage{amsfonts}
\usepackage{amsmath}
\usepackage{amssymb}
\usepackage{amstext}
\usepackage{amsthm}

\usepackage{booktabs}
\usepackage{dirtytalk}
\usepackage{enumitem}
\usepackage{hyperref}
\usepackage{mathtools}
\usepackage{microtype}
\usepackage{graphicx}
\usepackage{subfigure}
\usepackage{booktabs}
\usepackage[T1]{fontenc}
\usepackage{graphicx}

\usepackage{hyperref}
\usepackage[capitalize]{cleveref}

\usepackage{setspace}
\usepackage{subfigure}
\usepackage{wrapfig}

\usepackage{thmtools}
\usepackage{thm-restate}

\usepackage{nicefrac}
\usepackage{microtype}      
\usepackage[dvipsnames]{xcolor} 
\usepackage[parfill]{parskip}

\definecolor{salmon}{RGB}{234,153,153}
\definecolor{cornflowerblue}{RGB}{6,69,173}
\hypersetup{
    colorlinks,
    linkcolor={cornflowerblue},
    citecolor={cornflowerblue},
    urlcolor={salmon}
}

\theoremstyle{plain}

\newtheorem{theorem}{Theorem}[section]
\newtheorem{proposition}[theorem]{Proposition}
\newtheorem{lemma}[theorem]{Lemma}
\newtheorem{corollary}[theorem]{Corollary}
\theoremstyle{definition}
\newtheorem{definition}[theorem]{Definition}
\newtheorem{assumption}[theorem]{Assumption}
\theoremstyle{remark}

\newcommand{\absl}[1]{\left \lvert #1 \right \rvert}
\newcommand{\algo}{$\mbox{\textsc{Fast}}_{\mbox{\textsc{omp}}}$}
\newcommand{\bigo}[1]{\mathcal{O}\left ( #1 \right )}
\newcommand{\blimes}{\textsc{Isk}}

\newcommand{\dash}{\textsc{Dash}}
\newcommand{\dprod}[2]{\langle #1 , #2 \rangle}
\newcommand{\expect}[2]{\mathbb{E}_{#1}  \left[ \ #2 \ \right]}
\newcommand{\fast}{\textsc{Fast}}
\newcommand{\falgo}{f(\mathsf{S}^*)}
\newcommand{\fopt}{\mathtt{OPT}}
\newcommand{\greedy}{\textsc{Sds$_{\mbox{ma}}$}}
\newcommand{\ind}{\mathcal{I}}
\newcommand{\indspan}[1]{\mbox{Cond}(#1)}
\newcommand{\ignore}[1]{}
\newcommand{\lasso}{\mbox{Lasso}}

\newcommand{\omp}{\textsc{Sds$_{\mbox{omp}}$}}
\newcommand{\oracle}[1]{l ( #1 )}
\newcommand{\pnorm}[2]{\| #1 \|_{#2}}
\newcommand{\pr}[2]{\mathbb{P}_{#1} ( #2 )}
\newcommand{\random}{\mbox{\mbox{Random}}}
\newcommand{\rndseq}{\mbox{\textsc{rndSeq}}}

\newcommand{\shuff}{\mbox{\textsc{rndSeq}}\xspace}
\newcommand{\shuffle}[1]{\shuff\left ( #1 \right )}

\DeclareMathOperator*{\argmax}{arg\,max}

\title{\textbf{Fast Feature Selection with Fairness Constraints}}

\date{}

\usepackage{authblk}

\author[1]{Francesco Quinzan}
\author[2]{Rajiv Khanna}
\author[3]{Moshik Hershcovitch}
\author[4]{\\Sarel Cohen}
\author[3]{Daniel G. Waddington}
\author[4]{Tobias Friedrich}
\author[5]{Michael W. Mahoney}

\affil[1]{KTH Royal Institute of Technology}
\affil[2]{Purdue University}
\affil[3]{IBM Research}
\affil[4]{Hasso Plattner Institute}
\affil[5]{ICSI and UC Berkeley}

\begin{document}

\maketitle

\begin{abstract}
\noindent We study the fundamental problem of selecting \emph{optimal} features for model construction. This problem is computationally challenging on large datasets, even with the use of greedy algorithm variants. 
To address this challenge, we extend the adaptive query model, recently proposed for the greedy forward selection for submodular functions, to the faster paradigm of Orthogonal Matching Pursuit for non-submodular functions. The proposed algorithm achieves exponentially fast parallel run time in the adaptive query model, scaling much better than prior work.
Furthermore, our extension allows the use of downward-closed constraints, which can be used to encode certain fairness criteria into the feature selection process. We prove strong approximation guarantees for the algorithm based on standard assumptions. 
These guarantees are applicable to many parametric models, including Generalized Linear Models. 
Finally, we demonstrate empirically that the proposed algorithm competes favorably with state-of-the-art techniques for feature selection, on real-world and synthetic datasets.
\end{abstract}

\section{Introduction}
We study the fundamental problem of selecting a few features out of many for a given modeling problem, while satisfying additional side constraints. An an application, we also study how this setup can be used to encode certain notions of fairness in feature selection in a principled way.%
\footnote{In this paper, we consider several fairness constraints proposed in the literature, but there may be other notions of fairness which do not fall within our theoretical framework.}
Formally, given a function $l:\mathbb{R}^n \rightarrow \mathbb{R}_{>0}$ expressing the goodness of fit, we search for a set of features $\mathsf{S}$ maximizing the function
\begin{equation}
\label{eq:def_f}
    f(\mathsf{S}) \coloneqq \oracle{\boldsymbol \beta^{(\mathsf{S})} } - \oracle{\mathbf{0} }.
\end{equation}
Here, $\mathbf{0}$ represents the $n$-dimensional zero vector, and $\boldsymbol \beta^{(\mathsf{S})}$ is a vector maximizing $l(\cdot) $ with support in $\mathsf{S}$. If we denote by $\ind{}$ the set of all acceptable solution sets that satisfy the side constraints, then the feature selection optimization problem with side constraints can be formalized as
\begin{equation}
\label{eq:problem}
    \argmax_{\mathsf{S}\subseteq [n] \colon \mathsf{S} \in \ind{}} f(\mathsf{S}) = \argmax_{\mathsf{S}\subseteq [n] \colon \mathsf{S} \in \ind{}} \oracle{\boldsymbol \beta^{(\mathsf{S})} } - \oracle{\mathbf{0} },
\end{equation}
where $[n]$ is the index set of all the features. Often, $\ind{}$ corresponds to an $r$-sparsity constraint, i.e., a solution $\mathsf{S}$ is feasible if it contains at most $r$ features. Several algorithms have been proposed for feature selection under sparsity constraints. Some examples include forward step-wise selection methods \citep{DBLP:journals/corr/ElenbergKDN16,DBLP:conf/nips/QianS19}, the Orthogonal Matching Pursuit \citep{DBLP:journals/cacm/NeedellT10,DBLP:journals/corr/ElenbergKDN16,DBLP:conf/aistats/Sakaue20a}, forward-backward methods \citep{DBLP:conf/nips/JalaliJR11,DBLP:conf/icml/0002YF14}, Pareto optimization \citep{DBLP:conf/nips/QianYZ15,DBLP:conf/aaai/0001BF20}, Exponential Screening \citep{article_rigollet}, and gradient-based methods \citep{DBLP:conf/nips/0002TK14,DBLP:journals/jmlr/YuanLZ17}. The aforementioned algorithms, however, are computationally inefficient on large datasets. Furthermore, there are a limited number of studies that take into account additional side concerns (e.g. by encoding them as matroids~\citep{chen18b,gatmiry2018}), which can be crucial when deploying machine learning systems in the real world.

Recently, there has been a growing effort towards developing fair algorithms for many fundamental problems, such as regression and classification \citep{DBLP:conf/aistats/ZafarVGG17,DBLP:conf/nips/KimRR18,DBLP:conf/kdd/FeldmanFMSV15,DBLP:conf/icml/AgarwalBD0W18,DBLP:conf/aaai/Grgic-HlacaZGW18}, matching \citep{DBLP:conf/aistats/Chierichetti0LV19}, and summarization \citep{halabiMNTT20}. Several definitions of fairness can be incorporated in the learning process as additional side constraints \citep{DBLP:conf/aaai/Grgic-HlacaZGW18,DBLP:conf/www/Grgic-HlacaRGW18,GrgicHlaca2016TheCF,DBLP:conf/aistats/Chierichetti0LV19,halabiMNTT20,DBLP:conf/aistats/ZafarVGG17,DBLP:conf/colt/WoodworthGOS17,DBLP:conf/nips/DoniniOBSP18,DBLP:conf/icml/AgarwalBD0W18,DBLP:conf/icml/AgarwalDW19}. Motivated by this line of research, we consider the following  question: \emph{How can we efficiently perform feature selection, while taking into account additional constraints such as fairness?}

We address this question by proposing a novel algorithm that combines the paradigms of matching pursuit and the adaptive query model for the problem \eqref{eq:problem}. This algorithm is much faster than previously known techniques. At the same time, it allows incorporation of certain notions of fairness in the learning process, via a reduction to the $p$-system side constraints (see \citet{jenkyns1976efficacy} and Section~\ref{sec:fairness_constraints}).

We recognize that quantifying the full meaning of the notion of fairness with its societal context, as understood by human beings, may be hard to achieve through mathematical formalism alone~\citep{Selbst2019FairnessAA}. Indeed, $p$-systems may not explicate all current or future codifications of fairness. Our claim is not to \emph{solve} the problem of fairness itself, but rather to provide a theoretically sound framework for some of its currently accepted manifestations, with the hope of serving as a blueprint for further developments along similar lines.

Optimization problems as in~\eqref{eq:problem} under general side constraints are computationally challenging in practice. A major bottleneck lies in the evaluation of $\boldsymbol \beta^{(\mathsf{S})}$ for a given set $\mathsf{S}$, since this operation requires to re-train the model onto every candidate set $\mathsf{S}$. Another challenge is enforcing the side constraint $\ind{}$ that encodes the selection criteria, except in the trivial cases where the protected classes are known a priori~\citep{DBLP:conf/kdd/BeutelCDQWWHZHC19,DBLP:conf/nips/LahotiBCLPT0C20}. Our proposed algorithm is efficient with respect to these challenges.

\paragraph{Our Contribution.} We propose a novel matching pursuit algorithm for the constrained feature selection problem \eqref{eq:problem}. This algorithm uses oracle access to the gradient $\nabla l (\boldsymbol \beta^{(\mathsf{S})})$, and to an oracle for the evaluation of the feasibility of an input solution set. The feasibility oracle is well-aligned with previous related work \citep{DBLP:journals/corr/ElenbergKDN16,DBLP:conf/aistats/Sakaue20a}.

Our algorithm is based on a general technique called adaptive sequencing, that was recently proposed for submodular functions~\citep{DBLP:conf/stoc/BalkanskiRS19,DBLP:conf/icml/BreuerBS20}. However, previously proposed adaptive sequencing algorithms fail on problem \eqref{eq:problem}, as shown in Appendix~\ref{sec:motivating_example}. On the other hand, our algorithm converges after poly-logarithmic rounds in the adaptive query model. In each round, calls to the oracle functions can be performed in parallel, resulting in a dramatic speed-up.

The main technical contributions of this paper are two-fold: (i) we extend the adaptive query model to a class of non-submodular objectives using the paradigm of gradient based pursuit algorithms; (ii) we incorporate general downward-closed constraints in the optimization process beyond the standard sparsity constraints. Our main result can be stated as follows.
\begin{theorem}[informal] Denote with $\fopt$ the global maximum of the function $f(\cdot)$ as in \eqref{eq:problem}. There exists a randomized algorithm that outputs a set of features $\mathsf{S}^*$ such~that
\begin{equation*}
\frac{\mathbb{E}[f(\mathsf{S}^*)]}{\fopt} \geq \frac{1}{1 + p}\left (1 - \exp \left \{- \alpha (1 - \varepsilon)^2  \right \} \right ),
\end{equation*}
for tolerance $0 < \varepsilon < 1 $. Here, $p$ is a parameter that depends on $\ind{}$, and $\alpha$ is a parameter that depends on $l(\cdot)$. For $n$ total number of features, this algorithm uses $\bigo{\varepsilon^{-2} \log n}$ rounds of calls to  $\nabla l (\boldsymbol \beta^{(\mathsf{S})})$. Furthermore, this algorithm uses expected $\bigo{\varepsilon^{-2}\sqrt{r}\log n}$ rounds of calls to the oracle for the feasibility of an input solution set, where $r$ is the size of the largest feasible solution.
\end{theorem}
To the best of our knowledge, our algorithm is the fastest known algorithm for the general setting of maximizing non-submodular functions with side constraints, with provable guarantees and strong empirical performance (see Section~\ref{sec:experiments}). \citet{DBLP:conf/nips/QianS19} propose another algorithm for maximizing non-submodular functions that converges after poly-logarithmic rounds. However, their algorithm cannot handle general side constraints $\ind{}$ beyond sparsity by design. Hence, the algorithm of \citet{DBLP:conf/nips/QianS19} is unsuitable for more complex applications such as learning with fairness constraints~\citep{DBLP:conf/aaai/Grgic-HlacaZGW18}. Furthermore, for the standard $r$-sparsity constraint, we improve upon their approximation guarantee (see Theorem \ref{thm:approx}).

\paragraph{Technical Overview.} In our analysis, we face two major technical challenges. The first challenge is that of re-purposing the adaptive sequencing for functions that are not submodular, without a significant loss in the approximation guarantee. Adaptive sequencing has been so far employed only for maximizing submodular functions. Interestingly, it is known that standard adaptive sampling techniques do not guarantee constant factor approximation for functions with weak diminishing returns, which are typically invoked for feature selection theoretical studies~\citep{DBLP:journals/corr/ElenbergKDN16}.

The second challenge consists of integrating a constrained selection process based on orthogonal projections in the above adaptive sequencing framework. Common objective functions for feature selection do not have certain desirable properties (e.g., an antitone gradient) and the standard analysis for adaptive sequencing fails in our setting. To resolve these issues, we build upon the work of~\citet{DBLP:journals/corr/ElenbergKDN16} to establish a connection between gradient evaluations of functions that are restricted strong concave, and their marginal contributions to the optimization cost. This connection allows us to bound the gradient evaluations in terms of a discrete function, which is in turn used in the analysis to obtain the desired approximation guarantees. See Theorem~\ref{thm:approx} for the formal result and Appendix~\ref{appendix:A} for the proof.
\section{Preliminaries}
\emph{Notation.} We denote with $n$ the number of features, i.e., the dimension of the domain of $l(\cdot)$, and we define $[n] \coloneqq \{1, 2, \dots, n \}$. For any $s \in [n]$, we denote with $\mathbf{e}_s$ the unit vector, with a $1$ for the coefficient indexed by $s$, and $0$ otherwise. Feature sets are represented by sans script fonts, i.e., $\mathsf{S}, \mathsf{T}$. Vectors are represented by lower-case bold letters as $\boldsymbol{x}$, $\boldsymbol{y}$, $\boldsymbol \beta$ and matrices are represented by upper-case bold letters, i.e., $\boldsymbol{X}$, $\boldsymbol{Y}$, $\boldsymbol \Sigma$. For a feature set $\mathsf{S}$, we denote with $\boldsymbol \beta^{(\mathsf{S})}$ a vector maximizing $l(\cdot) $ with non-zero entries indexed by the set $\mathsf{S}$. For a feature set $\mathsf{T}$ and a parameter vector $\boldsymbol \beta$, we define $\nabla l(\boldsymbol \beta)_{\mathsf{T}} \coloneqq \dprod{\nabla l(\boldsymbol \beta)}{\sum_{s\in \mathsf{T}}\mathbf{e}_s}$. We denote with $\fopt$ the optimal value attained by the function $f(\cdot) $ as in \eqref{eq:problem}. We denote with $\ind{}$ the $p$-system side constraint, and with $r$ its rank, as defined in Section \ref{sec:problem_formulation}. The notation $\indspan{\mathsf{T}}$ denotes the set $\{ s \in [n]\setminus \mathsf{T}\colon \mathsf{T}\cup \{s\} \in \ind{} \}$.
\subsection{Problem Formulation}
\label{sec:problem_formulation}
We study optimization tasks as in the problem \eqref{eq:problem} under some additional assumptions on $l(\cdot)$, which are often satisfied in practical applications (see Appendix \ref{app:feature_selection}). Define an $r$-sparse subdomain as a set of the form $\Omega_r \coloneqq \{(\boldsymbol{x}, \boldsymbol{y}) \in \mathbb{R}^n \times \mathbb{R}^n \colon 
\pnorm{\boldsymbol{x}}{0} \leq r, 
\pnorm{\boldsymbol{y}}{0} \leq r, 
\pnorm{\boldsymbol{x}-\boldsymbol{y}}{0} \leq r\}$. We now define the notions of Restricted Strong Concavity (RSC) and Restricted Smoothness (RSM) of a function.

\begin{definition}[RSC, RSM~\citep{article_statistical_science}]
\label{def:RSC/RSM}
A function $l(\cdot)$ is said to be restricted strong concave (RSC) with parameter $m $ and restricted smooth (RSM) with parameter $M $ on a subdomain $\Omega_r$ iff, for all $(\boldsymbol{x}, \boldsymbol{y}) \in \Omega_r $ it holds that $ - \frac{m }{2} \pnorm{\boldsymbol{y} - \boldsymbol{x}}{2} \geq l(\boldsymbol{y}) - l(\boldsymbol{x})  - \dprod{\nabla l(\boldsymbol{x})}{\boldsymbol{y} - \boldsymbol{x}} \geq  - \frac{M }{2} \pnorm{\boldsymbol{y} - \boldsymbol{x}}{2}$. 
\end{definition}
We say that $l$ is $(M , m )$-(smooth, restricted concave), if it fulfills the conditions as in Definition \ref{def:RSC/RSM} with parameters $M$ and $m$. RSC/RSM often hold in practice, we refer the reader to Appendix \ref{app:feature_selection} for further discussion of these properties.

We model the side constraints as $p$-systems. In order to give an axiomatic definition of $p$-systems, we introduce additional terminology. Given a collection of feasible solutions $\ind{}$ over a ground set $[n]$ and a set $\mathsf{T} \subseteq [n]$, we denote with $\ind{}\mid_T$, the restricted feasible solution set, as the collection consisting of all sets $\mathsf{S} \subseteq \mathsf{T}$ s.t. $\mathsf{S} \in \ind{}$. We define $\indspan{\mathsf{T}}$ as the set of all $s \in [n]\setminus \mathsf{T}$ such that $\mathsf{T}\cup \{s\}\in \ind{}$.  A set $\mathsf{T}$ is a maximal independent set if it holds that $\indspan{\mathsf{T}} = \emptyset$ A base for $\ind{}$ is any maximal set $\mathsf{T} \in \ind{}$.

\begin{definition}[$p$-Systems \citep{jenkyns1976efficacy}]
\label{p-system}
A $p$-system $\ind{}$ over $[n]$ is a collection of subsets of $[n]$ such~that: (i) $\emptyset \in \ind{}$; (ii) for any two sets $\mathsf{S} \subseteq \mathsf{T} \subseteq [n]$, if $\mathsf{T} \in \ind{}$ then $\mathsf{S} \in \ind{}$; (iii) for any set $\mathsf{T} \subseteq [n]$ and any bases $\mathsf{S}, \mathsf{U} \in \ind{}\mid_{\mathsf{T}}$ it holds $\absl{\mathsf{S}} \leq p \absl{\mathsf{U}}$.
\end{definition}
The second defining axiom is commonly referred to as subset-closure or downward-closed property. The rank $r$ of a $p$-system $\ind{}$ is defined as the maximum cardinality of any feasible solution $\mathsf{T}\in \ind{}$.

Armed with these definitions, we can re-visit the problem~\ref{eq:problem} with additional assumptions, where the set of feasible solutions $\mathcal{I}$ is a $p$-system, and  $l(\cdot)$ is restricted strong concave and smooth. Our problem formulation is a strict generalization of previous related works \citep{DBLP:journals/corr/ElenbergKDN16,DBLP:conf/aistats/Sakaue20a,DBLP:conf/aistats/Chierichetti0LV19} which considered the $r$-sparsity constraints encoded as $\ind{}\coloneqq \{ \mathsf{T}\subseteq [n] \colon \absl{\mathsf{T}} \leq r \}$ which is a special case of the more general $p$-system constraints. Further, $p$-systems are also a strict generalization of the matroid-type constraints considered in submodular literature~\citep{chen18b,gatmiry2018}.
\subsection{Embedding Fairness via \texorpdfstring{$p$}{}-Systems}
\label{sec:fairness_constraints}
In our framework, the $p$-system $\ind{}$ enumerates which sets of features are considered \say{fair}. That is, a set of features $\mathsf{S}$ is acceptable as fair if and only if $\mathsf{S} \in \ind{}$. Our framework is very flexible, and can handle a large variety of constraints, including many constraints used to enumerate notions of fairness (see, e.g., Section 4 by \citet{DBLP:conf/aistats/Chierichetti0LV19} and Section 2 by \citet{DBLP:conf/aaai/Grgic-HlacaZGW18}). Non-trivial examples of $p$-systems side constraints are also found in the context of data summarization \citep{DBLP:conf/icml/MirzasoleimanBK16,DBLP:conf/colt/FeldmanHK17,DBLP:conf/aaai/MirzasoleimanJ018,DBLP:conf/aistats/QuinzanD0021}. We now describe how some additional well-established notions of fairness can also be embedded as $p$-systems.

\paragraph{Procedural fairness metrics:} Procedural fairness focuses on selecting features based on perceived notion of fairness as envisioned by human beings during the process of decision making, rather than on the fairness of the outcome. It is measured by gauging ``the
degree to which people consider various features to be fair"~\citep{GrgicHlaca2016TheCF}. This is in contrast to measuring fairness of the \emph{outcomes} of such decisions, for example,
by down weighing decisions that affect users of protected groups (e.g., race,
gender). 

 In this work, we consider measures for procedural fairness studied by \citet{GrgicHlaca2016TheCF} and \citet{DBLP:conf/aaai/Grgic-HlacaZGW18}. However, our framework is not specific to these definitions. These measures consist of monotone set functions $h: 2^{[n]}\rightarrow [0,1]$. For an input feature set $\mathsf{T} \subseteq [n]$, the value $h(\mathsf{T})$ quantifies the perceived fairness of $\mathsf{T}$, with $h(\mathsf{T}) = 0$ corresponding to maximum fairness and $h(\mathsf{T}) = 1$ corresponding to maximum unfairness. We can take a specific example, where $h(\cdot)$ is the \emph{feature-apriori} unfairness~\citet{GrgicHlaca2016TheCF}. For a given feature $s \in [n]$, denote with $\mathcal{U}_s$ the set of users that perceive a feature to be fair. For a set of features $\mathsf{T}\subseteq [n]$,~\citet{GrgicHlaca2016TheCF} define the feature-apriori unfairness as
\begin{equation*}
    h
    (\mathsf{T}) \coloneqq 1 - \frac{\absl{\bigcap_{s \in \mathsf{T}}\mathcal{U}_s}}{\absl{\mathcal{U}}}.
\end{equation*} 
Sets of features $\mathsf{T}\subseteq [n]$ are selected only if the value $h(\mathsf{T})$ is below a certain threshold. Given a monotone set function $h$ as described above, we can enumerate \say{fair} sets of features as $ \ind{}_{\mbox{acc}}^{\lambda}\coloneqq \left \{ \mathsf{T}\in [n]\colon h(\mathsf{T}) \leq \lambda \right \}  $. Here, $\lambda \in [0, 1]$ is a user-defined parameter, which determines the trade-off between fairness and accuracy. Since, we require $h(\cdot)$ to be monotone, the $\ind{}_{\mbox{acc}}^{\lambda}$ satisfies the downward closed property and is a $p$-system (see Definition~\ref{p-system}). The similar notions of feature-disparity fairness and feature-accuracy fairness can be embedded as $p$-systems in a similar fashion. 
While procedural fairness may not imply fairness of the outcome, it has been observed that in some cases procedurally fair feature sets maintain good outcome fairness \citep{GrgicHlaca2016TheCF,DBLP:conf/aaai/Grgic-HlacaZGW18}.

\paragraph{Feature partitions:}
Our proposed approach also includes as a special case the framework proposed by \citet{DBLP:conf/icml/CelisKS0KV18}. Here, features are grouped into disjoint clusters $[n] = \mathsf{X}_1 \cup \dots \cup \mathsf{X}_\ell$. The constraints are specified using $\lambda_j$ which encodes the maximum number of features that can be selected from cluster $\mathsf{X}_j$. In other words, features $\mathsf{T}\subseteq [n]$ is then feasible if the number of data-points intersecting a class $\mathsf{X}_j$ does not exceed the corresponding threshold $\lambda_j$. Formally, we define the set of constraints $ \ind{}_{\mbox{cl}}^{\boldsymbol \lambda} \coloneqq \left \{ \mathsf{T}\subseteq [n]\colon \absl{\mathsf{T}\cap \mathsf{X}_j} \leq \lambda_j \ \forall j \in [\ell] \right \}  $. This set of constraints is a \emph{matroid}, which is a $p$-system with $p = 1$.

A generalization of the aforementioned partition matroid was considered by~\citet{halabiMNTT20}.
For each element in any partition set $\mathsf{X}_j$, we are given a lower- and an upper-bound on the number of elements that can be selected from this set. Bounds are denoted by $\ell_j$ and $u_j$ respectively. The set of constraints can be written as $\ind{}_{\mbox{p}} \coloneqq \left \{ \mathsf{T}\subseteq [n]\colon \ell_j \leq\absl{\mathsf{T}\cap \mathsf{X}_j} \leq c_j \ \forall j \in [\ell] \right \}$.
This set of constraints is, in general, not a $p$-system. However,~\citet{halabiMNTT20} show that one can consider a relaxation of the constraint set $\overline{\ind{}}_{\mbox{p}} \coloneqq \left \{ \mathsf{T}\subseteq [n]\colon \mathsf{T} \subseteq \mathsf{S} \ \text{for any set} \ \mathsf{S} \in \ind{}_{\mbox{p}}\right \} $, which is equivalent from the optimization perspective.  Any monotone optimization objective (as in \eqref{eq:def_f}) yields the same solution sets on $\ind{}_{\mbox{p}}$ and $\overline{\ind{}}_{\mbox{p}}$. The set of constraints $\overline{\ind{}}_{\mbox{p}}$ is a \emph{matroid}, i.e., a $p$-system with $p = 1$~\citep{DBLP:conf/aussois/Edmonds01}.
\subsection{The Computational Model}
We assume access to an oracle that returns $\nabla l (\boldsymbol \beta^{(\mathsf{T})})$, for a given input set $\mathsf{T}$. \citet{DBLP:journals/corr/ElenbergKDN16} highlight the benefits of using this oracle model for feature selection, since access to the gradient is available from the inner optimization. In the case of a linear model, the gradient $\nabla l (\boldsymbol \beta^{(\mathsf{T})})$ can be easily estimated for various metrics $l$ expressing the goodness of fit. For instance, if $l$ is the log-likelihood function, then the gradient can be computed in explicit form. For more complex models, stochastic lower-bounds of $\log \pr{\boldsymbol \beta ^{(\mathsf{S})}}{\boldsymbol{x}}$ can be used, and then differentiated \citep{,DBLP:conf/nips/BamlerZOM17,DBLP:conf/iclr/Nowozin18}. Similar considerations hold for other metrics, such as the $R^2$ objective \citep{DBLP:journals/corr/ElenbergKDN16}.

We also assume access to the independence oracle of the underlying $p$-system $\ind{}$. The independence oracle takes as input a set $\mathsf{T}$, and returns as output a Boolean value, true if $\mathsf{T}\in \ind{}$ and false otherwise. This oracle is often assumed for optimizing functions over $p$-systems \citep{DBLP:conf/icml/MirzasoleimanBK16,DBLP:conf/aistats/QuinzanD0021}. Our method also works assuming access to a rank oracle, or a span oracle. We refer the reader to \citep{DBLP:conf/stoc/ChekuriQ19} for a description of these oracle models.

We evaluate performance using the notion of adaptivity. The adaptivity refers to the number of sequential rounds of the algorithm, wherein polynomial number of parallel queries are made in each round~\citep{DBLP:conf/stoc/BalkanskiS18}. Formally, given an oracle $f$, an algorithm is $r$-adaptive if every query $q$ to the oracle $f$ occurs at a round $i \in [r]$ such that $q$ is independent of the answers $f(q')$ to all other queries $q'$ at round $i$. This notion is closely related to the Parallel Random Access Machines (PRAM) model, as shown in Appendix \ref{appendix:PRAM}. We evaluate empirical speedup by the adaptivity of the oracle to evaluate $\nabla l (\boldsymbol \beta^{(\mathsf{T})})$. We also evaluate the adaptivity of the independence oracle for the $p$-system $\ind{}$.
\section{Algorithmic Overview}
\label{sec:algo_overview}
\begin{algorithm}[t]
	\caption{\algo{}}
	\label{alg}
	$\mathsf{S} \gets \emptyset$;\\
	\While{the number of iterations is less than $\varepsilon^{-1}$ and $\indspan{\mathsf{S}} \neq \emptyset$}{
	$\mathsf{X} \gets \{ s \in [n]\colon \{ s\}\cup \mathsf{S} \in \ind{}\}$;\\
	$t \gets (1 - \varepsilon)\frac{m}{\absl{\mathsf{T}}M}\pnorm{\nabla l (\boldsymbol \beta ^{(\mathsf{S})})_{\mathsf{T}}}{2}^2$ with $\mathsf{T} \subseteq \mathsf{X}$ maximizing $\pnorm{\nabla l (\boldsymbol \beta ^{(\mathsf{S})})_{\mathsf{T}}}{2}^2$ s.t. $ \absl{\mathsf{T}} \leq r $;\\
    \While{$\mathsf{X} \neq \emptyset$ and $\indspan{\mathsf{S}} \neq \emptyset$}{
    $\{a_1, a_2, \dots, a_k\} \gets \shuffle{\mathsf{X}, \mathsf{S}}$ and define $\mathsf{S}_j \gets \mathsf{S} \cup \{a_1, \dots, a_j\}$;\\
    \textbf{observe} $\mathsf{X}_j \gets \{s \in \mathsf{X} \colon \dprod{ \nabla \oracle {\boldsymbol \beta^{(\mathsf{S}_{j})}}}{ \mathbf{e}_s }^2 \geq t \mbox{ and } s \in \indspan{\mathsf{S}_j} \}$;\\
    $j^* \ \gets \min_j \{ j \colon \lvert  \mathsf{X}_j \rvert < (1 - \varepsilon) \absl{\mathsf{X}} \}$;\\
    $\mathsf{X} \gets \mathsf{X}_{j^*}$ and $\mathsf{S} \gets \mathsf{S}_{j^*}$;\\
    }
    }
   	\textbf{return} $\mathsf{S}$;
\end{algorithm}
Our algorithm, which we call \algo{}, is presented in Algorithm \ref{alg}. This algorithm is based on a technique called adaptive sequencing \citep{DBLP:conf/stoc/BalkanskiRS19,DBLP:conf/icml/BreuerBS20}, which was recently proposed for highly scalable maximization of submodular functions.

Say $\mathsf{X}$ is the complete set of candidate features, and $\mathsf{S}$ is the current solution. Starting from $\mathsf{S} \gets \emptyset$, the \algo{} iteratively generates a random sequence of features $\{a_1, a_2, \dots, a_k\}$ with the \shuff{} sub-routine (details in Appendix~\ref{appendix:randSeq}), such that the set $\{a_1, a_2, \dots, a_k\}\cup \mathsf{S}$ is a maximal independent set of $\ind{}$. After a sequence is generated, the \algo{} identifies a prefix $\{a_1, \dots, a_{j^*}\}$ that is added to the current solution. The index $j^*$ defining this prefix is chosen such that it holds $\lvert  \mathsf{X}_{j} \rvert \geq (1 - \varepsilon) \absl{\mathsf{X}}$, for all $0 \leq j < j^*$. This inequality ensures that any point added to the current solution yields $ \dprod{ \nabla \oracle{ \boldsymbol\beta^{(\mathsf{S})}}}{ \mathbf{e}_s }^2  \geq t $ in expected value. Finally, the ground set $\mathsf{X}$ is updated as to include only those points that yield a good improvement to the new solution. The $\shuff$ sub-routine used to generate $\{a_1, a_2, \dots, a_k\}$ corresponds to Algorithm A by \citet{DBLP:journals/jcss/KarpUW88}. Here, $k$ is the size of the independent set returned in the current iteration. 

\paragraph{Adaptive sequencing via matching projections.} Our proposed algorithm differs from previous adaptive sequencing techniques, since it does not use queries to the function $f $ as defined in \eqref{eq:def_f}. Instead, our algorithm uses oracle access to the function $\nabla l(\boldsymbol  \beta^{( \mathsf{S})})$, and features $s \in [n]$ are added to the current solution if it holds $\dprod{\nabla l(\boldsymbol \beta ^{(\mathsf{S}_j)})}{\mathbf{e}_s}\geq t$ in expected value, with $t$ a threshold updated during run time. As such, our approach extends the applicability of adaptive sequencing to gradient-based pursuit methods, as opposed to the standard value-oracle based methods in earlier works. Finding optimal solutions with this technique requires much less computation than the previous approach \citep{DBLP:conf/stoc/BalkanskiRS19}, by which points $s$ are selected if it holds $f(\mathsf{S}_j\cup \{s\}) \geq t$. This is because it is usually much faster to compute an inner product, than evaluating $f(\cdot)$ for every candidate $\mathsf{S}$.

\paragraph{Implicit estimates of $\fopt$.} All algorithms based on adaptive sampling techniques proposed so far require an estimate of $\fopt$, the value of the optimal solution set. This value is typically not known a priori. To circumvent this problem, previous algorithms perform multiple runs for various guesses of $\fopt$ \citep{DBLP:conf/icml/BreuerBS20}, or they use additional preprocessing steps \citep{DBLP:conf/soda/FahrbachMZ19}. Our algorithm has the significant advantage that $\fopt$ is estimated implicitly. For a function $l(\cdot)$ that is $(m,M)$-(smooth, restricted concave), we have that $\max_{\{ \mathsf{T}\colon \absl{\mathsf{T}} \leq k\}} \pnorm{\nabla l (\boldsymbol \beta ^{(\mathsf{S})})_{\mathsf{T}}}{2}^2 \geq 2m \left (\fopt - f(\mathsf{S}) \right )$. A proof of this result is deferred to Appendix \ref{appendix:A}, and it is based on \citet{DBLP:journals/corr/ElenbergKDN16}. Hence, the \algo{}  estimates $\fopt $ with a single oracle valuation, and it does not require multiple runs or preprocessing steps.

\paragraph{Finding the set $\mathsf{X}_{j^*}$.} Common optimization functions $l(\cdot)$ for feature selection lack certain desirable properties, such as an antitone gradient. Hence, in contrast to the submodular case, the sequence $\{\absl{\mathsf{X}_j}\}_j$ as in Line 7 of Algorithm~\ref{alg} is not monotonic. For this reason, it is not possible to estimate the index $j^*$ with a binary search, as other adaptive sequencing algorithms do \citep{DBLP:conf/icml/BreuerBS20}. We require $\bigo{r}$ phases of re-training to estimate $j^*$, whereas the general adaptive sequencing technique would require $\bigo{rn}$ phases of re-training \citep{DBLP:conf/stoc/BalkanskiRS19}, due to the different oracle model.
\section{Approximation Guarantees}
\label{sec:approx}\vspace{-1mm}
In this section, we study the approximation guarantees for Algorithm \ref{alg}, when solving Problem~\eqref{eq:problem}.
\begin{restatable}{theorem}{approx}
\label{thm:approx}
Define the support selection function $f(\cdot )$ as in~\eqref{eq:def_f}, for the given function $l(\cdot) $ that is (M,m)-(restricted smooth, restricted strong concave), on the sparse sub-domain $\Omega_{2r}$. Consider a $p$-system $\ind{}$ of rank $r$ over $[n]$, and let $\mathsf{S}^*$ be the  output of Algorithm~\ref{alg} while $\fopt$ is the optimum solution set for the Problem \ref{eq:problem}. Then,
\begin{equation*}
\frac{\expect{}{\falgo}}{\fopt} \geq \frac{1}{1 + p}\left (1 - \exp \left \{-(1 - \varepsilon)^2 \frac{m^3}{M^3}\right \} \right ) ,
\end{equation*}
for all $0 < \varepsilon < 1 $. Furthermore, in the specific case when $\ind{}$ is $r$-sparsity constraint over $[n]$, then,
\begin{equation*}
\frac{\expect{}{\falgo}}{\fopt} \geq \left (1 - \exp \left \{-(1 - \varepsilon)^2 \frac{m^2}{M^2}\right \} \right ).
\end{equation*}
\end{restatable}
\begin{figure}[t]
\centering
\subfigure[]{\label{fig:sparsity:k_time}\includegraphics[width=0.3\textwidth]{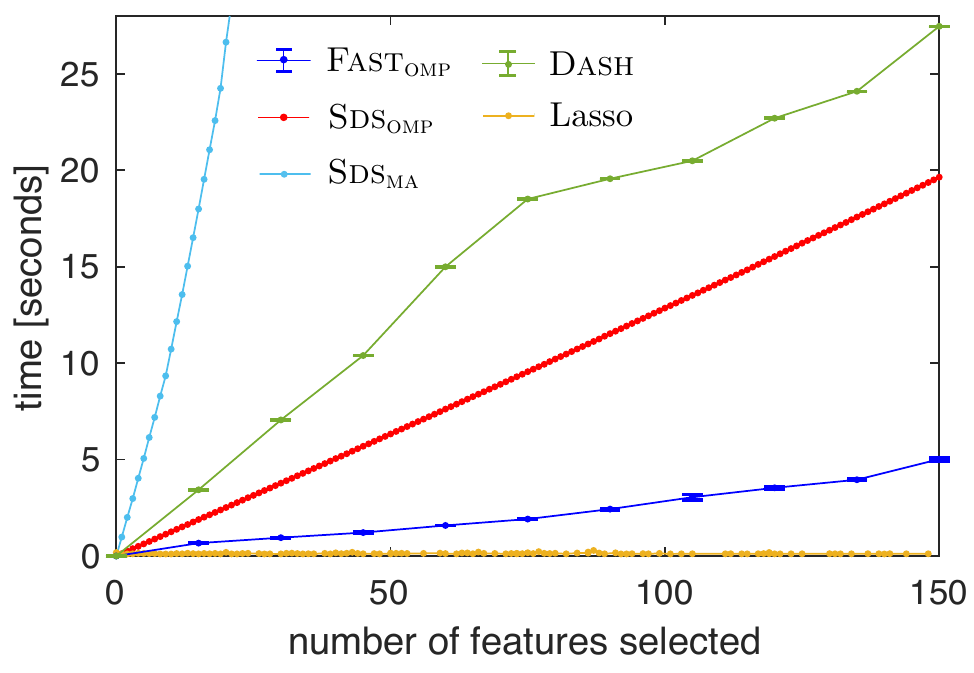}}
\subfigure[]{\label{fig:sparsity:time_metric}\includegraphics[width=0.3\textwidth]{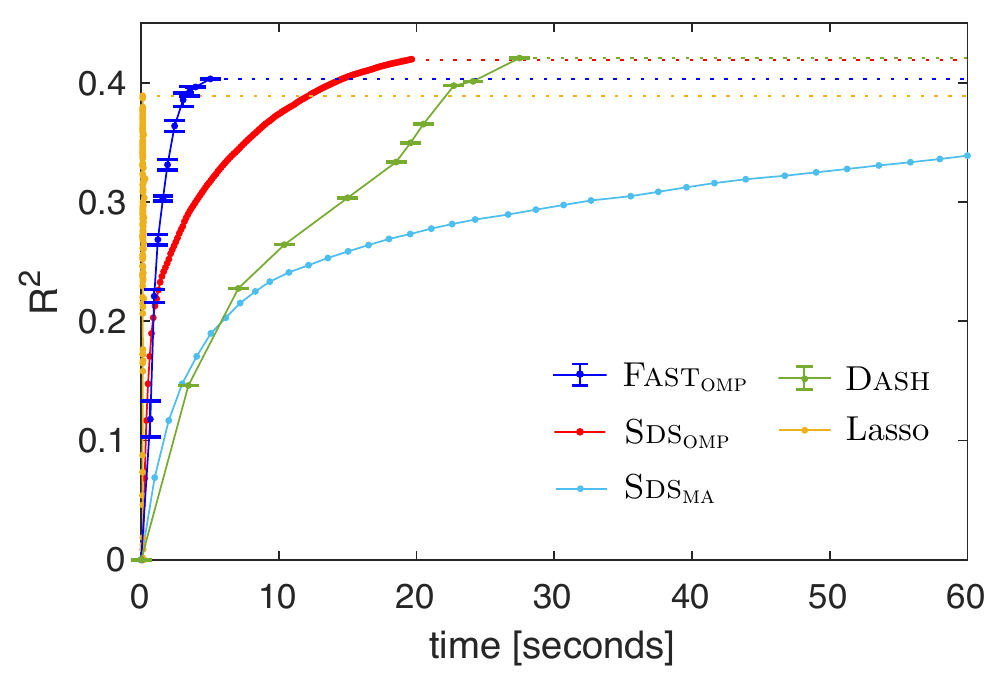}}
\subfigure[]{\label{fig:sparsity:k_score}\includegraphics[width=0.3\textwidth]{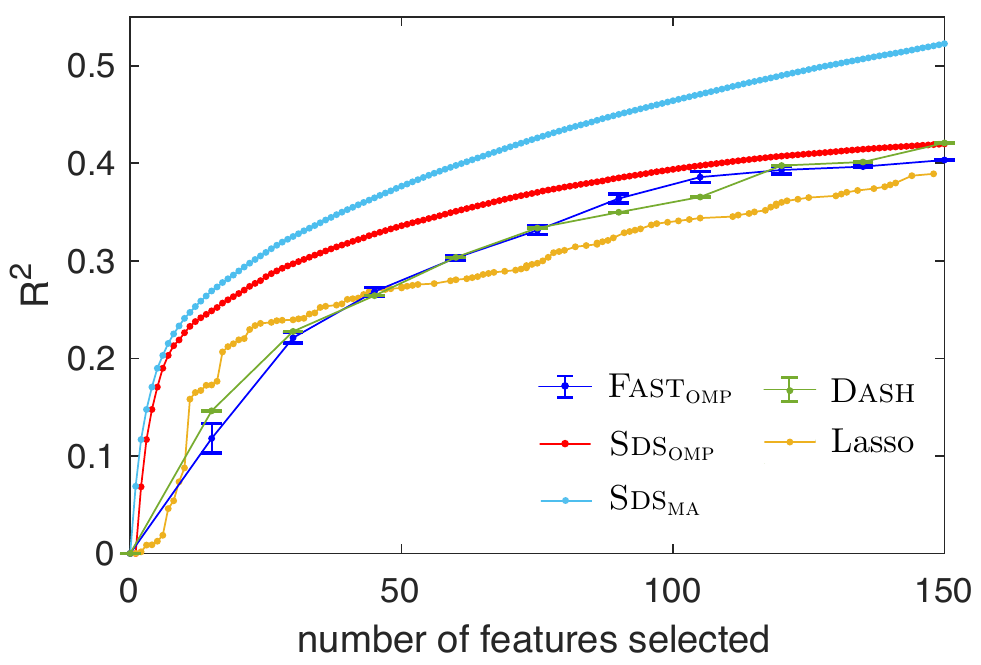}}
\subfigure[]{\label{fig:synth:k_time}\includegraphics[width=0.3\textwidth]{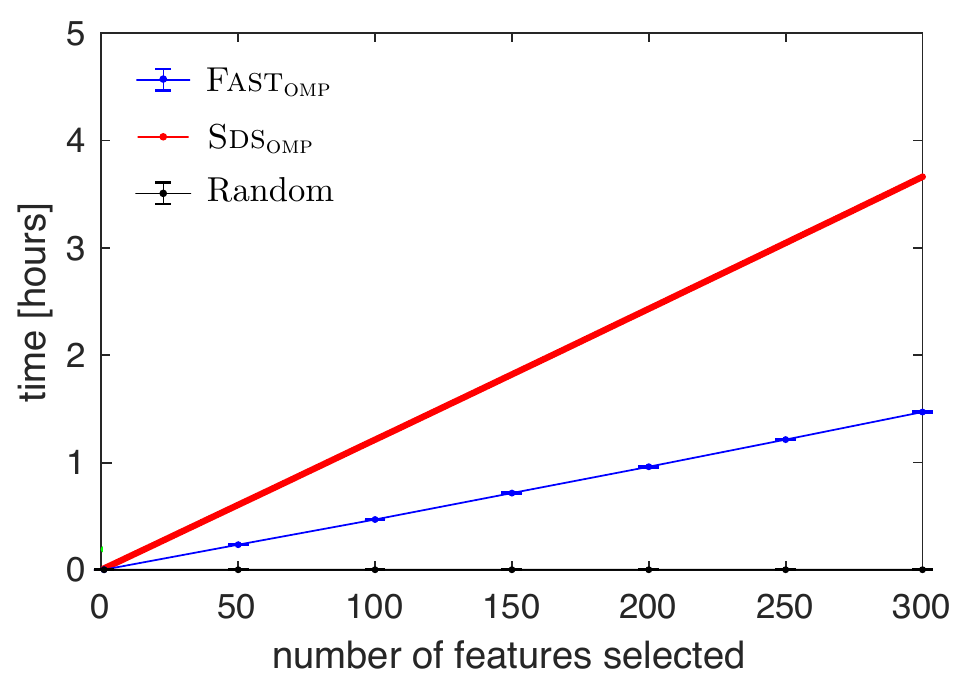}}
\subfigure[]{\label{fig:synth:time_metric}\includegraphics[width=0.3\textwidth]{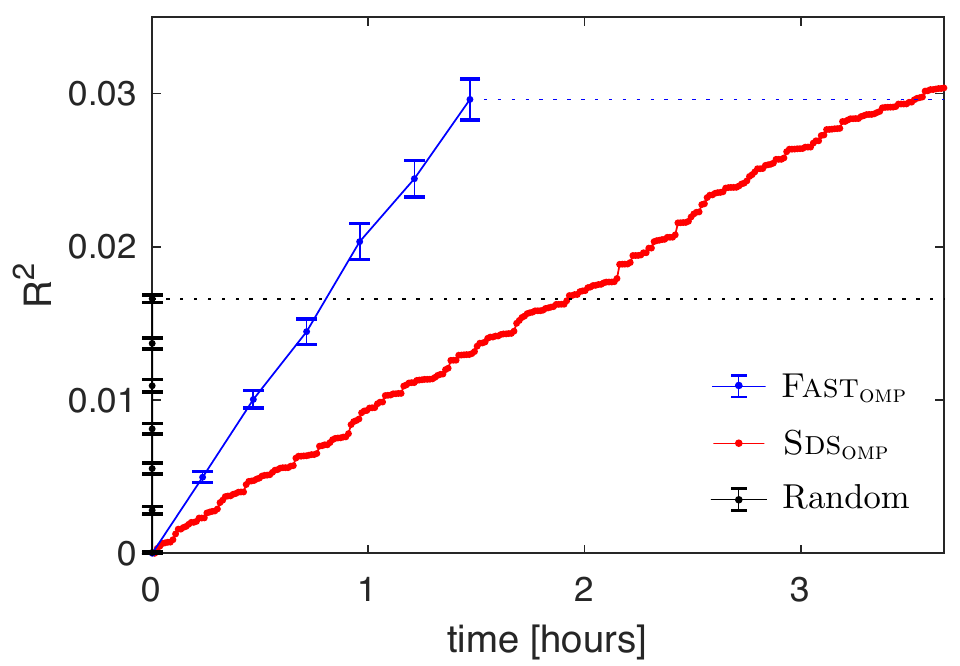}}
\subfigure[]{\label{fig:synth:k_score}\includegraphics[width=0.3\textwidth]{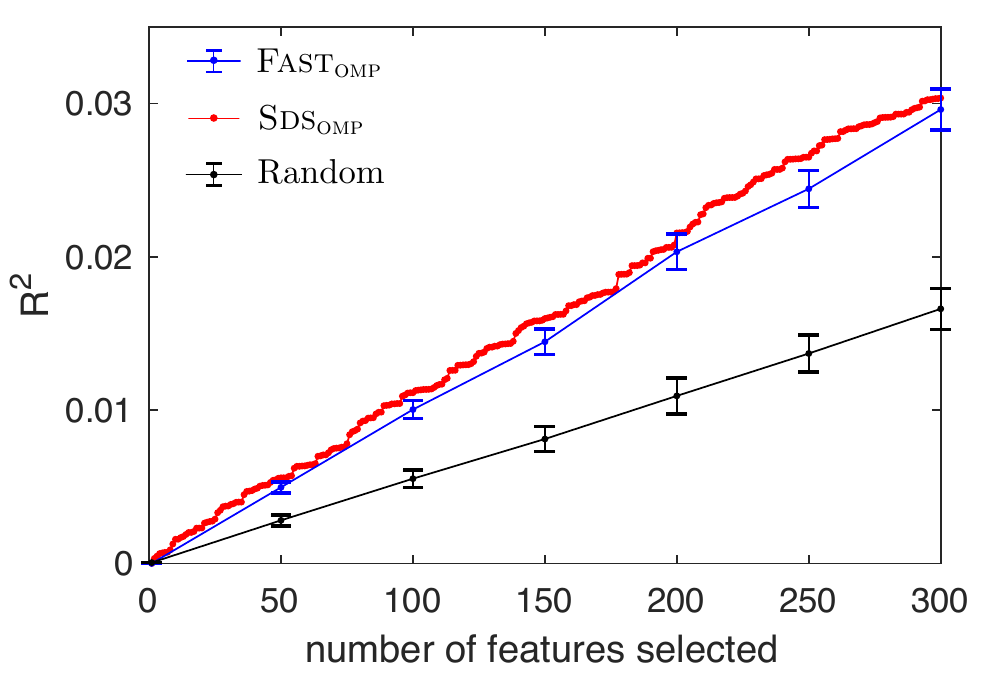}}
\subfigure[]{\label{fig:real:k_time}\includegraphics[width=0.3\textwidth]{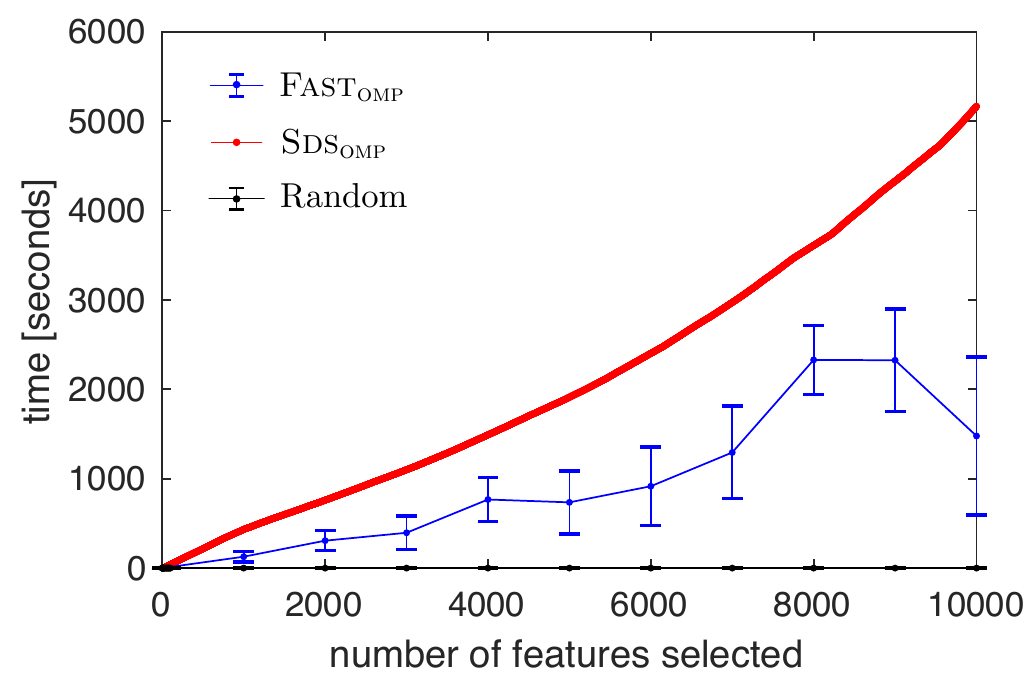}}
\subfigure[]{\label{fig:real:time_metric}\includegraphics[width=0.3\textwidth]{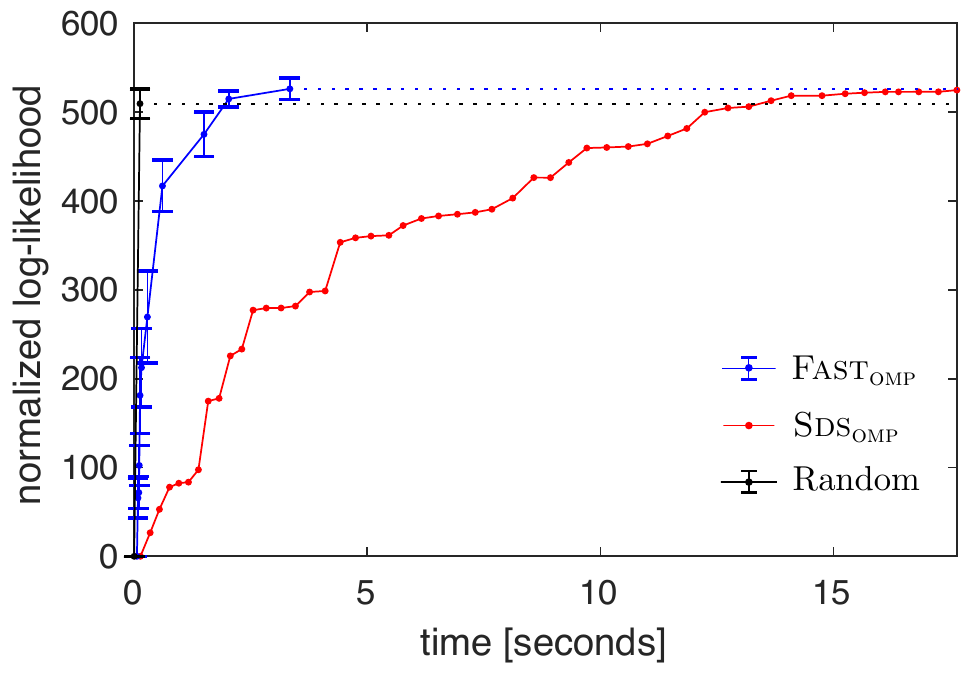}}
\subfigure[]{\label{fig:real:k_score}\includegraphics[width=0.3\textwidth]{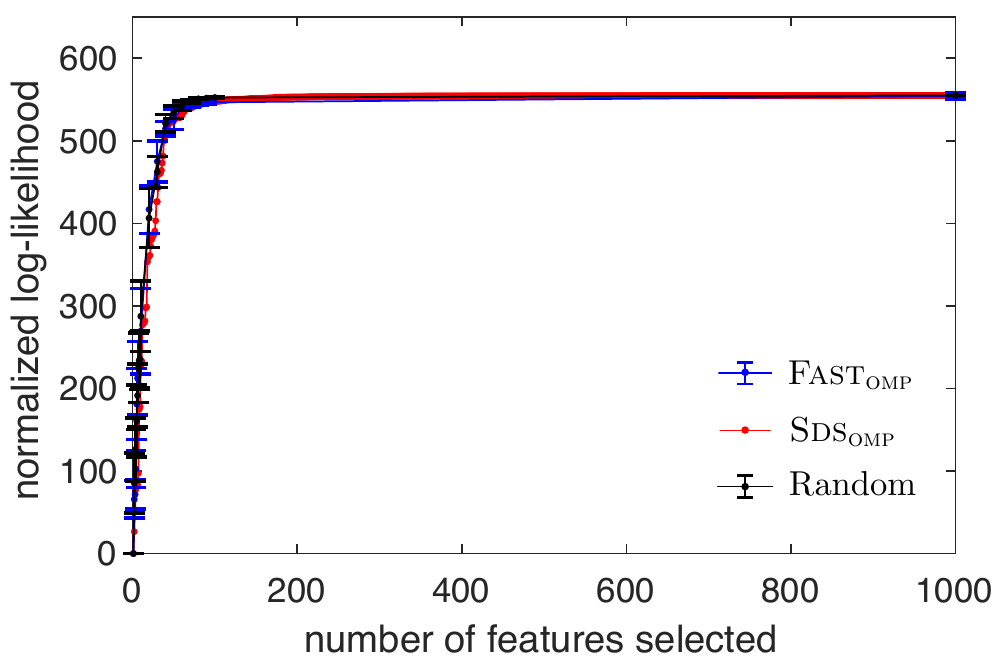}}
\caption{Results on the Synthetic Unconstrained Dataset (top row), the Synthetic Dataset with Constraints (mid row), and the Pan-Cancer Dataset (bottom row), as in Section \ref{sec:experiments}.}
\label{fig:synth}\label{fig:sparsity}
\end{figure}
A full proof of this theorem is deferred to Appendix \ref{appendix:A}. We remark that, if $\ind{}$ is a $r$-sparsity constraint, then the approximation guarantee of Theorem \ref{thm:approx} is asymptotically better than the guarantee attained by other parallel algorithms for this problem, such as the \dash{} \citep{DBLP:conf/nips/QianS19}. Specifically, as proven in Theorem 1 by \citet{DBLP:conf/nips/QianS19}, the \dash{} yields an approximation of $1 - \exp \{m^4 /M^4 \}  - \varepsilon $ on this problem. Furthermore, the \dash{} cannot handle general $p$-system side constraints.

The parameter $p$ in Theorem \ref{thm:approx} is always upper-bounded by the maximum number of features $r$ that we wish to select for model construction. This upper-bound still holds if one assumes additional underlying fairness constraints. Additional assumptions on the $p$-system may yield an improved bound on $p$. Finally, previous related work \citep{halabiMNTT20} reduces some fairness constraints to a matroid, in which case our analysis applies with $p = 1$. We also provide bounds for the run time of the \algo{} as follows.
\begin{restatable}{theorem}{runtime}
\label{thm:run_time}
Algorithm \ref{alg} terminates after $\bigo{\varepsilon^{-2} \log n}$ rounds of calls to the oracle function, and it uses at most $\bigo{\varepsilon^{-2}r \log n}$ oracle queries. Furthermore, Algorithm \ref{alg} requires expected $\bigo{\varepsilon^{-2}\sqrt{r}\log n}$ independent calls to the oracle for the $p$-system $\ind{}$, and the total expected number of calls to the oracle for the $p$-system $\ind{}$ is $\bigo{\varepsilon^{-2} n r \log n}$.
\end{restatable}
The proof of Theorem \ref{thm:run_time} is deferred to Appendix \ref{appendix:adptivity_run_time}. 
The estimates on the rounds of adaptivity extends to the \textsc{pram} model. If we denote with $d_l $ the depth required to evaluate the oracle function on a set, then the \algo{} has $\bigo{\varepsilon^{-2} d_l \log n}$ depth. Note that the rounds of independent calls to the oracle are sub-linear, but not poly-logarithmic in the problem size. The reason is that the \shuff{} sub-routine requires expected $\bigo{\sqrt{n}}$ rounds of independent calls to the oracle for the $p$-system (see Appendix \ref{appendix:randSeq}).

Some authors have proposed non-adaptive techniques for feature selection. If $\ind{}$ is an $r$-sparsity constraint, then \citet{DBLP:journals/corr/ElenbergKDN16,DBLP:conf/aistats/Sakaue20a} provide an algorithm that require $\bigo{1}$ sequential oracle calls. This algorithm selects $r$ best points $s \in [n]$ independently, according to the values $f(\{s\})$, or the value of the inner product $\langle \nabla l(\mathbf{0}), \mathbf{e}_s \rangle$. However, oblivious feature selection techniques for general $p$-system side constraints require $\Omega(r)$ sequential calls to the independence oracle. These techniques are impractical on large dataset, when the feasibility of a solution set is computationally expensive.
\section{Experiments} 
\label{sec:experiments}
In this section, we present empirical evidence of the efficacy of the proposed algorithm. We provide extensive experiments on both synthetic and real world datasets. All experiments are performed on Python 3 on a server that runs Linux with Intel Xeon E5-2630 v4 with 40 

\begin{wrapfigure}{l}{0.48\textwidth}
  \begin{center}
  \includegraphics[width=0.4\textwidth]{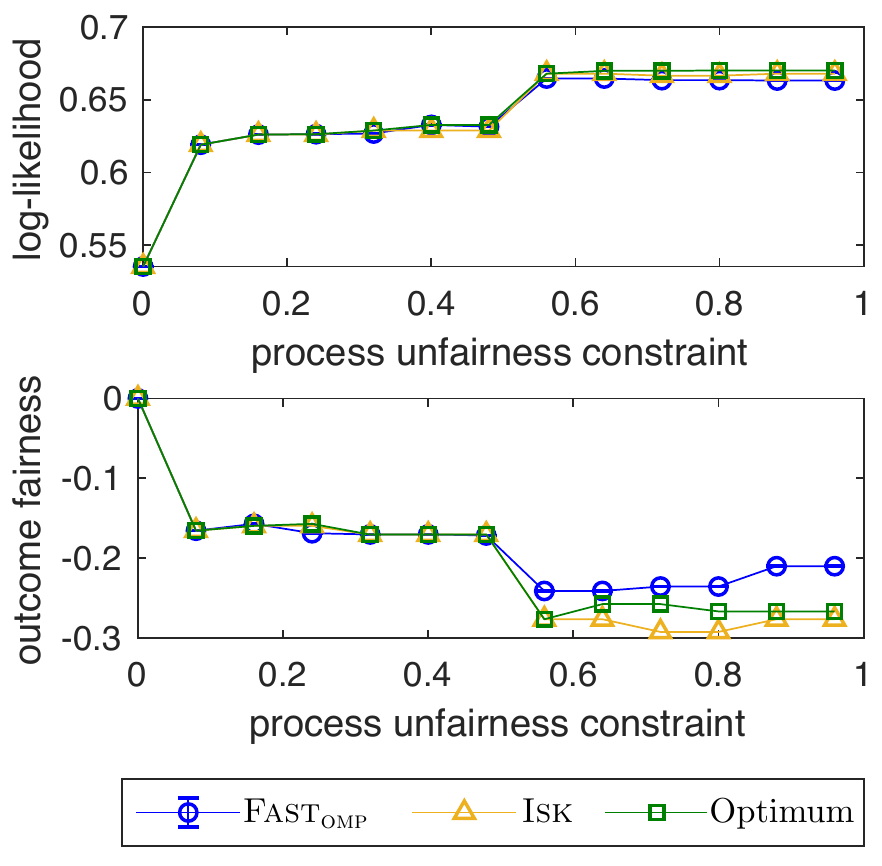}
  \caption{Experiments on the COMPAS Dataset, as detailed in Section~\ref{sec:experiments}. For the $\textsc{Fast}_\textsc{omp}$ we perform multiple runs and we report on the sample mean. On this dataset, the variance in accuracy and outcome fairness is negligible for $\textsc{Fast}_\textsc{omp}$. \label{fig:fairness}}
  \end{center}
  \vspace{-25pt}
\end{wrapfigure}
CPUs at 2.2GHz. We also tested our algorithms on an emerging CPU-attached persistent memory technology using MCAS \citep{MCAS1,MCAS_ADO} (see Appendix \ref{appendix:mcas}). We now describe the datasets we used for our experiments.

\paragraph{Synthetic Datasets.}  We generate two synthetic datasets of different sizes. The first dataset has $500$ features, $1000$ observations, and it has size $\sim 10$MB; the other dataset has $10^6$ features, $10^4$ observations, and it has size $\sim 19$GB. In both cases, features are sampled with a joint Gaussian distribution with mean vector $\boldsymbol \mu = \mathbf{0}$, and a covariance matrix $\boldsymbol \Sigma$ designed to ensure that $10\%$ of the features are highly correlated with the response, and the remaining features have low correlation with the response variable. We then add posterior uniform noise, and normalize the observations.

\paragraph{The CGARN Pan-Cancer Dataset.} This dataset has about $2 \times  10^4$ features and $804$ observations, with size $\sim 201$MB \citep{563846}. In this dataset, the features embed information on the genome sequences of $802$ patients affected with cancer, and the observations consists of a pseudo-Boolean array, with $1$ if the corresponding patient has PRAD cancer and $0$ otherwise.

\paragraph{The ProPublica COMPAS DataSet.} We consider the well-known ProPublica COMPAS dataset, which is a pretrial risk assessment instrument \citep{COMPAS_DATASET}. It was constructed in 2016, using data of defendants from Broward County, FL, who had been arrested in 2013 or 2014. This dataset was intended to be used to predict if a criminal was likely to re-offend, based on previous arrest charges and demographic information. Predictions based on the COMPAS datasets were found to be racially biased \citep{doi:10.1177/0049124118782533123,doi:10.1177/0049124118782533} (see Appendix \ref{appendix:pro_publica} for details).

We use the COMPAS dataset to reproduce the experiments by \citet{DBLP:conf/aaai/Grgic-HlacaZGW18}. These experiments use the COMPAS dataset to predict if a defendant faces risks of recidivism, and study the trade-off between fairness and accuracy achieved by the feature-apriori accuracy, feature-accuracy fairness, and feature-disparity fairness (see Section \ref{sec:fairness_constraints}). Our interest in using the COMPAS dataset is only to quantitatively compare our algorithm with previous related work. We do not claim that normalizing the task of recidivism prediction is something that can be made \say{fair} with this approach.

\subsection{Benchmarks.}
In this section, we describe the benchmark algorithms we use for comparison against the proposed framework. Our goal in empirical evaluation is to illustrate the accuracy vs speedup trade-off that follows from our computational model of adaptive sampling based matching pursuit under constraints. The main purpose of using a technique like adaptive sampling is to obtain speedups by reducing the number of oracle evaluations compared to their non-adaptive counterparts, with some admissible loss in accuracy~\citep{DBLP:conf/stoc/BalkanskiRS19}.  Indeed, there are many other feature selection algorithms. We choose feature selection benchmarks that help us illustrate the said speedup obtained when using our method while ensuring graceful degradation in accuracy in simulated and real-world use-cases.
These algorithms include popular selection methods \citep{DBLP:conf/icml/KalimerisKS19,DBLP:conf/icml/KrauseC10,DBLP:journals/corr/ElenbergKDN16,DBLP:conf/aaai/Grgic-HlacaZGW18}. Other popular algorithms for feature selection include the Maximum Relevance Minimum Redundancy (mRMR) \citep{DBLP:conf/dsaa/ZhaoAW19,DBLP:journals/jbcb/DingP05} and the Conditional Mutual Information Maximisation (CMIM) filters \citep{DBLP:journals/jmlr/Torkkola03} among others. These algorithms iteratively adds features by maximizing suitable objectives, such as scores based on the F-statistic, or Mutual Information gain. However, these algorithms for general $p$-system side constraints require $\Omega(r)$ sequential calls to the independence oracle for a solution size of $r$. Thus, such methods are impractical and will be trivially too slow compared to our method. We consider the following algorithms to compare against:

\begin{enumerate}[label={$\bullet$},itemsep=0pt,topsep=0pt]
\item \textbf{\greedy{}:} Starting from the empty set, this algorithm adds feasible points to the current solution in a greedy fashion \citep{DBLP:conf/icml/KrauseC10,DBLP:journals/corr/ElenbergKDN16}. This algorithm uses oracle access to the function $f $ as in \eqref{eq:def_f}. 
\item \textbf{\omp{}:} Starting from the empty set, this algorithm iteratively adds a feature $s$ to the current solution $\mathsf{S}$ if it maximizes the dot product $\dprod{\nabla l(\boldsymbol \beta ^{(\mathsf{S})})}{\mathbf{e}_s}$ \citep{DBLP:conf/icml/KrauseC10,DBLP:journals/corr/ElenbergKDN16}.
\item \textbf{\dash{}:} This algorithm follows a computationally similar model, and achieves strong approximation guarantees on the subset selection problem, under the RSC/RSM assumption \citep{DBLP:conf/nips/QianS19}. It also uses oracle access to the function $f$ but can only handle $r$-sparsity constraints.
\item \textbf{\blimes{}:} This algorithm is the iterated submodular-cost knapsack algorithm proposed by \citet{DBLP:conf/nips/IyerB13}. It was used by \citet{DBLP:conf/aaai/Grgic-HlacaZGW18}  to perform feature selection on the ProPublica COMPAS dataset. The \blimes{} can only handle $r$-sparsity constraints, and it has no known guarantees for the problem \eqref{eq:problem}.
\item \textbf{\lasso{}:} We also compare against the popular \lasso{} regression. Tuning \lasso{} to obtain a solution of a desired size is hard \citep{DBLP:conf/icml/MairalY12}. In our experiments, we vary the parameter manually and benchmark against the resulting solution size.
\item \textbf{\random{}:} This simple algorithm outputs a maximum independent set of $\ind{}$ chosen uniformly at random. We use the \rndseq{} to generate this set (see Appendix \ref{appendix:randSeq}).
\end{enumerate}
\subsection{Results}
In this section, we present extensive empirical evidence that validates our theory that the proposed algorithm \algo{} is significantly faster and more scalable than the baselines, while maintaining competitive performance in accuracy.

\paragraph{Unconstrained Synthetic Dataset.} We consider a sparse linear regression task on the small synthetic dataset ($\sim10$MB), where the goal is to search for a set of few features maximizing the $R^2$ objective. Figure \ref{fig:sparsity:k_time} showcases the run time of each algorithm for a fixed number of features selected. We observe that our algorithm significantly outperforms baselines. In Figure \ref{fig:sparsity:time_metric}, we fix an upper-bound of $k = 150$ on the number of features selected, and we compare the solution accuracy of each algorithm, for a given time budget. We observe that the \algo{} outperforms the non-oblivious algorithms. Furthermore, we observe that the \greedy{} yields significantly worse performance than baselines. In Figure \ref{fig:sparsity:k_score} we display the solution quality versus the number of features selected. We observe that the \algo{}, the \omp{}, the \dash{}, the \lasso{} achieve similar solution quality. The \greedy{} yields best performance, but it is much slower than the other algorithms.

\paragraph{Synthetic Dataset with Constraints.}
We search for a set of features maximizing the $R^2$ objective, for the large synthetic dataset ($\sim19$GB). We consider a randomly generated $p$-system on top of the features. We do not report on the results for the \greedy{}, because it was too slow on account of the dataset being too large. We do not test the \dash{}, the \blimes{}, and the \lasso{} since they cannot handle $p$-system side constraints by design.

To illustrate scalability on such a large dataset, we compare speed of feature selection while satisfying the side constraints (Figure \ref{fig:synth:k_time}) and the corresponding solution accuracy with respect to time spent (Figure \ref{fig:synth:time_metric}) for selecting upto $k=300$ features. We observe that our algorithm \algo{} is significantly faster while maintaining a good quality solution when compared to the \omp{} and the \random{}. In Figure \ref{fig:synth:k_score}, we further observe that the \algo{} and the \omp{} achieve similar solution quality for a given number of features.
 
\paragraph{CGARN Pan-Cancer Dataset} We use the logistic regression to predict if patients have PRAD cancer, or other types of cancer. We enforce a randomly generated $p$-system on the features. We search for features maximizing the normalized log-likelihood. Again, \greedy{} was too slow on this dataset and we do not report on the results for it. We also do not test the \dash{}, the \blimes{}, and the \lasso{} since they cannot handle $p$-systems by design.

To illustrate the scalability and speed of \algo{}, we show that it is much faster in selecting a given number of features (Figure \ref{fig:real:k_time}) and achieves much better accuracy for a given run time (\ref{fig:real:time_metric}) compared to the baselines. Figure~\ref{fig:real:time_metric} is presented till selection of $k=50$ features, since adding more features did not improve the metric by much as seen in Figure~\ref{fig:real:k_score} which shows that the  algorithms \algo{} and \omp{} perform similarly in terms of the log-likelihood for a given number of selected features.

 
\paragraph{ProPublica COMPAS Dataset.} In this section, we reproduce the experiments by \citet{DBLP:conf/aaai/Grgic-HlacaZGW18} on the COMPAS dataset. We use regularized logistic regression to predict the recidivism risk and use fairness constraints given by the feature-apriori accuracy \citep{DBLP:conf/aaai/Grgic-HlacaZGW18}. These constraints are encoded as $p$-system side constraints $\ind{}_{\mbox{acc}}^\lambda$ as described in Section \ref{sec:fairness_constraints}. We report on the outcome fairness of each output feature set, by estimating the racial bias of the corresponding classifier. Following \citet{DBLP:conf/aaai/Grgic-HlacaZGW18,DBLP:conf/innovations/KleinbergMR17,DBLP:conf/aistats/ZafarVGG17}, we examine the false positive (FPR) and false negative (FNR) rates for whites (w) and non-whites (nw) as 
$$\text{\emph{outcome fairness}} = - \absl{\mbox{FPR}_{w} - \mbox{FPR}_{nw}} - \absl{\mbox{FNR}_{w}- \mbox{FNR}_{nw}}.$$
This measure of fairness varies between $-2$ and $0$, with $-2$ corresponding to maximum unfairness and $0$ to maximum fairness.

The results for this set of experiments are displayed in Figure \ref{fig:fairness}, where we plot the accuracy and outcome fairness as a function of the parameter $\lambda$ for the constraints $\ind{}_{\mbox{acc}}^\lambda$. We compare the solution quality and fairness of the \algo{}, against the \blimes{}, and the solution with best possible accuracy (optimum). We observe that the \algo{} achieves nearly optimal solution. Although the outcome fairness inevitably decreases for increasing process unfairness, the \algo{} maintains a more graceful degradation than the other algorithms.
\section{Conclusion and Ethics Discussion}
\label{conclusion}\vspace{-2mm}
We have extended the adaptive sequencing framework, first proposed for maximizing submodular functions, to the setting of feature selection, via the matching pursuit paradigm. Our analysis yields strong performance and approximation guarantees, which are better than previously known results. Furthermore, our proposed formulation is more general than previously considered, as it can handle $p$-system constraints. While our main contributions are theoretical and algorithmic, we apply these results for fair feature selection. Our framework ensures approximation guarantees, as long as the constraints can be encoded as $p$-systems. 

A major hurdle in further research into fair learning is the lack of gold standard benchmarks. The COMPAS dataset is used for empirical evaluations in several studies  However, its use for benchmarking has also been criticized~\citep{bao2021its}. Furthermore, over-tuning notions of fairness for a single dataset could be problematic.

Fairness criteria should also take into account the contextual grounding of the dataset, and the trained model that operates within. As such, the mathematical formalism of fairness as constraints may evolve. We hope that our framework motivates further research into addressing the above concerns about fairness in an algorithmically scalable way.
\bibliographystyle{abbrvnat}
\bibliography{bibliography.bib}
\newpage
\onecolumn
\renewcommand{\thesection}{\Alph{section}}
\setcounter{section}{0}
\newpage
\onecolumn
\renewcommand{\thesection}{\Alph{section}}
\setcounter{section}{0}
\noindent {\LARGE\textbf{Appendix}}
\section{Motivating Example}
\label{sec:motivating_example}
In this section, we motivate our analysis by showing that the adaptive sequencing prototype by \citet{DBLP:conf/stoc/BalkanskiRS19} does not work for feature selection. this example can also be used to show that similar algorithms, such as \textsc{fast} \citep{DBLP:conf/icml/BreuerBS20} do not work for feature selection.

The adaptive sequencing prototype is presented in Algorithm \ref{fast}. We refer to this algorithm as the \fast{}. Starting from the empty set, the \fast{} generates at every iteration a uniformly random sequence $\{a_1, \dots, a_k \}$ of the elements $\mathsf{X}$ not yet discarded. This sequence can be sampled uniformly at random, or using the $\shuff{}$ sub-routine described in Appendix \ref{appendix:randSeq}. Afterwards, the \fast{} filters elements that have a high marginal contribution, when added to $\mathsf{S}$, and it determines the prefix of $\{a_1, \dots, a_k \}$ that is added to the current solution $\mathsf{S}$. This prefix has the property that there is a large fraction of elements in $\mathsf{X}$ with high contribution to the current solution $\mathsf{S}$.

Following and example provided by \citet{DBLP:journals/corr/ElenbergKDN16}, we show that the \fast{} fails on a simple linear regression task with three features. For a fixed parameter $z > 0$, consider the following variables:
\begin{align*}
    \mathbf{y} & = [1, 0, 0]^T \qquad \mathbf{x}_2 = [z, \sqrt{1 - z^2}, 0]^T \\
    \mathbf{x}_1 & = [0, 1, 0]^T
    \qquad \mathbf{x}_3 = [\delta z, 0, \sqrt{1 - \delta^2 z^2}]^T\\
\end{align*}
Note that all variables have unit norm. Our goal is to choose two of the three variables $\{\mathbf{x}_1, \mathbf{x}_2, \mathbf{x}_3 \}$ that best estimate $\mathbf{y}$, with respect to the $R^2$ objective. To this end, we introduce additional notation. For a given index set $\mathsf{S}\subseteq [3]$, we denote with $\mathbf{X}_{\mathsf{S}}$ the matrix whose columns consists of the features indexed by $\mathsf{S}$. For instance, for $\mathsf{S} = \{1,3 \}$ it holds
\begin{equation*}
  \mathbf{X}_{\{1,3 \}} =
  \left[ {\begin{array}{cc}
    0 & \delta z              \\
    1 & 0                     \\
    0 & \sqrt{1 - \delta^2 z^2} \\
  \end{array} } \right].
\end{equation*}
With this notation, we define the objective function for our problem as
\begin{equation}
\label{eq:r2}
    f(\mathsf{S}) = R^2(\boldsymbol \beta^{(\mathsf{S})} ) - R^2(\mathbf{0}) = (\mathbf{y}^T \mathbf{X}_{\mathsf{S}}) (\mathbf{X}_{\mathsf{S}}^T \mathbf{X}_{\mathsf{S}})^{-1} (\mathbf{X}_{\mathsf{S}}^T \mathbf{y}),
\end{equation}
with $R^2(\cdot)$ the $R^2$ objective evaluated on the model for an input parameter vector $\boldsymbol \beta^{(\mathsf{S})}$. Using this formula, one can easily see that it holds
\begin{align*}
& \\
    & f(\{1\}) = 0     \qquad \qquad \ \ \ f(\{1,2\}) = 1 \\
    & f(\{2\}) = z^2   \qquad \quad \ \ \ \ \ f(\{1,3\}) = \delta^2 z^2 \\
    & f(\{3\}) = \delta^2 z^2 \qquad \quad \ f(\{2,3\}) = (1 + \delta) z^2 + \delta^2 z^4\\
\end{align*}
Clearly, the optimal solution is $ f(\{1,2\})$. However, on this instance the \fast{} outputs the solution $\mathsf{S} = \emptyset$, attaining an $f$-value of $f(\mathsf{S}) = 0$. The following lemma holds.
\begin{algorithm}[t]
	\caption{\fast{}}
	\label{fast}
	$\mathsf{S} \gets \emptyset$, $t \gets \max_{e \in [n]} f(e)$;\\
	\For{$\Delta$ iterations }{
	$\mathsf{X} \gets [n] $;\\
    \While{$\mathsf{X} \neq \emptyset$}{
    $\{a_1, a_2, \dots, a_k\} \gets \shuffle{\mathsf{X}, \mathsf{S}}$;\\
    $X_i \gets \{e \in \mathsf{X} \colon f_{\mathsf{S} \cup \{a_1, \dots, a_i\}}(e) \geq t \ \text{and} \ \mathsf{S} \cup \{a_1, \dots, a_i, e\} \in \ind{} \}$\;
    $i^* \gets \min\{i \colon \absl{\mathsf{X}_i } \leq (1 - \varepsilon)\absl{\mathsf{X}} \} $;\\
    $\mathsf{S} \gets \mathsf{S} \cup \{a_1, \dots, a_i\}$;\\
    $\mathsf{X} = \mathsf{X}_{i^*}$
    }
    $t \gets (1 - \varepsilon ) t$;\\
    }
   	\textbf{return} $\mathsf{S}$;
\end{algorithm}
\begin{lemma}
Consider the \fast{} optimizing the function $f(\mathsf{S})$ as in \eqref{eq:r2}, over sets $\mathsf{S}\subseteq [3]$ of size at most $\absl{\mathsf{S}} \leq k$ with $k = 2$. Then, for any constant $\varepsilon > 0$, the \fast{} outputs either the solution $\{1, 3\}$ or the solution $\{2,3\}$ with probability at least $1 - (2/3)^{\varepsilon^{-1} \log 1/\delta}$. In particular, the expected approximation guarantee of \fast{} converges to zero, for $\delta, z \to 0$.
\end{lemma}
\begin{proof}
At the beginning of the optimization process, the constant $t$ is set to $t = \delta^2 z^2$ and a sequence $\{a_1, a_2\}$. This sequence yield $\{a_1\} = \{3\}$ at least with probability $1/3$. If $\{a_1\} = \{3\}$ sequence is sampled, then the point $\{3\}$ is added to the current solution. Otherwise, the value $t$ decreases of a multiplicative factor of $(1 - \varepsilon)$, and a new sequence is sampled. As long as $t > z^2$, the \fast{} can only output a solution that contains the point $\{3\}$. Hence, the solution $\{1, 2\}$ can only be sampled after $t$ decreases to a value $t < z$, which requires at least $\varepsilon^{-1} \log 1/\delta$ iterations of the outer loop. Hence, the probability of sampling the solution $\{1, 2\}$ can be upper-bounded by the probability that no sequence with $\{a_1\} = \{3\}$ is sampled during the first $\varepsilon^{-1} \log 1/\delta$ iterations of the outer-loop. This probability can be estimated as $2/3^{\varepsilon^{-1} \log 1/\delta}$. Hence, the \fast{} outputs either the solution $\{1, 3\}$ or the solution $\{2,3\}$ with probability at least $1 - (2/3)^{\varepsilon^{-1} \log 1/\delta}$.
\end{proof}
We remark that in this example, the \algo{} outputs the optimal solution in one iteration, at least with constant probability. We can bound the optimal parameters and in Line 4 of Algorithm 1 in a similar fashion as in Proposition \ref{cor:approx1}, and obtain that $m$ and $M$ are bounded as $m \geq 1 - \sqrt{1 - z^2}$ and $M \leq 1 + \sqrt{1 - z^2}$. It follows that the parameter in Line 4 of Algorithm 1 is upper-bounded as $t \leq (1 - \varepsilon) (1 - \sqrt{1 - z^2})/2 \leq z^2$, for $\delta$ sufficiently small. Suppose now that at the beginning of the iteration, a sequence $\{a_1, a_2\} = \{ 2, 1\}$. Since $t \leq z^2$, then the entire sequence is added to the current solution $\{2,1\}$ and the algorithm outputs the optimum. Note that the desired sequence is sampled with probability $1/6$. It follows that the \algo{} outputs the optimum at least with constant probability. Hence, the \algo{} maintains a constant-factor approximation guarantee.
\section{Weak Submodularity}
\label{sec:submoduarity}
Consider a function $l$ that is restricted smooth and restricted strong concave . It is well known that the corresponding function $f$ as in \eqref{eq:def_f} has weak diminishing return properties. These diminishing returns property is called \emph{weak submodularity}, and it was first introduced by \citet{DBLP:conf/icml/DasK11} to study statistical subset selection problems. Functions that exhibit weak submodularity, i.e., weakly submodular functions, are defined in terms of the submodularity ratio, as follows.
\begin{definition}[Weak submodularity, Definition 2.3 by \citet{DBLP:conf/icml/DasK11}]
Consider a function $f\colon 2^{[n]} \rightarrow \mathbb{R}_{\geq 0}$. The submodularity ratio is defined as the largest scalar $\gamma$ such that
\begin{equation*}
    \sum_{j \in \mathsf{S}}\left ( f(\mathsf{L}\cup \{j\}) - f(\mathsf{L}) \right ) \geq  \gamma \left ( f(\mathsf{L}\cup \mathsf{S}) - f(\mathsf{L}) \right ),
\end{equation*}
for all sets $\mathsf{L}, \mathsf{S}\in [n]$ such that $ \mathsf{L}\cap \mathsf{S} \neq \emptyset$.
We say that $f$ is $\gamma$- weakly submodular if its submodularity ratio is $\gamma$.
\end{definition}
There is a well-known connection between weak submodularity and Problem \ref{eq:problem}. This connection was discovered by \citet{DBLP:journals/corr/ElenbergKDN16}, and it can be formalized as follows.
\begin{theorem}[Theorem 1 by \citet{DBLP:journals/corr/ElenbergKDN16}]
Define $f$ as in \eqref{eq:def_f}, with a function $l$ that is $(m_{\absl{\mathsf{U}} + k}, M_{\absl{\mathsf{U}} + k})$-(strongly concave, smooth) on $\Omega_{\absl{\mathsf{U}}+ k}$, and $\tilde{M}_{\absl{\mathsf{U}}+1}$ smooth on $\Omega_{\absl{\mathsf{U}}+ 1}$. Fix a set $\mathsf{U}\in [n]$, and denote with $\gamma_{\mathsf{U},k}$ the largest scalar such that
\begin{equation*}
    \sum_{j \in \mathsf{S}}\left ( f(\mathsf{L}\cup \{j\}) - f(\mathsf{L}) \right ) \geq  \gamma_{\mathsf{U},k} \left ( f(\mathsf{L}\cup \mathsf{S}) - f(\mathsf{L}) \right ),
\end{equation*}
for all sets $\mathsf{L}, \mathsf{S}\in [n]$ such that $ \mathsf{L}\cap \mathsf{S} \neq \emptyset, \mathsf{L} \subseteq \mathsf{U}, \absl{\mathsf{S}} \leq k$. Then, the constant $\gamma_{\mathsf{U},k}$ is lower-bounded as
\begin{equation*}
    \gamma_{\absl{\mathsf{U}} , k} \geq \frac{m_{\absl{\mathsf{U}}  + k}}{\tilde{M}_{\absl{\mathsf{U}}  + 1}} \geq \frac{m_{\absl{\mathsf{U}}  + k}}{M_{\absl{\mathsf{U}} + k}}.
\end{equation*}
\end{theorem}
\section{Restricted Strong Concavity and Smoothness}
\label{app:feature_selection}
Given a set of observations and a parametric family of distributions $\{p(\ \cdot \ ; \boldsymbol \beta) \mid \boldsymbol \beta \in \Omega \}$ with $\Omega \subseteq \mathbb{R}^n$, we wish to identify a vector of parameters $\boldsymbol \beta $ maximizing the goodness of fit for these observation, according to a chosen measure $l $. For generalized linear models, common measures for feature selection are restricted strong concave and restricted smooth. We study the log-likelihood and the coefficient of determination, although analogous results hold for other similar statistics \citep{DBLP:conf/nips/QianS19,DBLP:conf/icml/DasK11}.
\paragraph{Maximizing the log-conditional.} Assuming that the response follows a distribution in an exponential family, the log-conditional can be written as
\begin{equation}
\label{eq:log_likelihood-function}
\log p(\mathbf{y} \mid \mathbf{X}; \boldsymbol \beta) = h^{-1} (\tau) - Z(\mathbf{X}, \boldsymbol \beta) + g(\mathbf{y}, \tau)
\end{equation}
with $Z $ the log-partition function, and $\tau$ the dispersion parameter. The log-conditional is commonly used for learning. In the case of the simple linear model, it is possible to derive approximation guarantees in terms of the eigenvalues of $\mathbf{X}$. Additional assumptions on the random design of $\mathbf{X}$ ensure that the log-conditional objective is restricted smooth and restricted strong concave. We refer the reader to Appendix \ref{appendix:B} for a discussion on these results. 
\paragraph{Maximizing the $R^2$ objective.}  Consider a linear model with a normalized response variable $\mathbf{y}$. The $R^2$ objective is defined as
\begin{equation}
\label{eq:coefficient_of_determination}
    R^2_{\mathbf{X}, \mathbf{y}}(\boldsymbol \beta) \coloneqq 1 - \expect{}{(\mathbf{y} - \dprod{\mathbf{X}}{\boldsymbol \beta})^2}. 
\end{equation}
This function is a popular measure for the goodness of fit. The function $ R^2_{\mathbf{X}, \mathbf{y}}(\boldsymbol \beta)$ is restricted strong concave and restricted smooth, with parameters depending on the properties of the matrix $\boldsymbol{X} $. A discussion on these results is deferred to Appendix \ref{appendix:proof_of_cor}.
\subsection{Restricted Strong Concavity and Smoothness of the Log-Conditional}
\label{appendix:B}
In this section we discuss results concerning the $(M,m)$-(smotheness, strong concavity) of the log-conditional for generalized linear models. Consider a feature matrix $\mathbf{X} \in \mathbb{R}^{n \times d} $ and response variable $\mathbf{y} $. Assuming that the response follows a distribution in an exponential family, the log-conditional can be written as
\begin{equation}
\label{eq:glm1AB}
\log p(\mathbf{y} \mid \mathbf{X}; \boldsymbol \beta) = h^{-1} (\tau) - Z(\mathbf{X}, \boldsymbol \beta) + g(\mathbf{y}, \tau)
\end{equation}
with $Z $ the log-partition function, and $\tau$ the dispersion parameter. We give approximation guarantees for the objective
\begin{equation}
\label{eq:glm1A}
l (\boldsymbol \beta) = \log p(\mathbf{y} \mid \mathbf{X}; \boldsymbol \beta) - \eta \pnorm{\boldsymbol \beta}{2}^2,
\end{equation}
for some parameter $\eta \geq 0$. We show that the function $l$ is restricted smooth and restricted strong concave under various assumptions. We start discussing the simplest case of a logistic regression. We then study the general case as in \eqref{eq:glm1AB}, under the assumption that $l$ has a non-zero regularization term. We then conclude with the general case, i.e., no assumptions on the regularization term.
\paragraph{Logistic regression.} The log-likelihood function for the logistic regression is defined as 
\begin{equation}
\label{objective:function2}
   l(\boldsymbol \beta^{(\mathsf{S})}) \coloneqq \sum_i y_i \dprod{\mathbf{X}_{\mathsf{S}}}{\boldsymbol \beta} - \log \left ( 1 - e^{\dprod{\mathbf{X}_{\mathsf{S}}}{\boldsymbol \beta}} \right ),
\end{equation}
with $\mathbf{X}_{\mathsf{S}}$ the matrix of all features indexed by $\mathsf{S}$ and $y_i$ the $i$-th observation, i.e., the $i$-th coefficient of the response $\boldsymbol{y}$.  For this class of log-likelihood functions, the following result holds.
\begin{proposition}
\label{cor:approx2}
The log-likelihood function $l$ for the logistic regression as in \eqref{objective:function2}, is $(m,M)$-(restricted smooth, restricted strong concave) with parameters
\begin{equation*}
    m \coloneqq \min_{\mathsf{S}} \lambda_{\min} (\mathbf{X}_{\mathsf{S}}^T \mathbf{X}_{\mathsf{S}}) \ \mbox{and} \ M \coloneqq \max_{\mathsf{S}} \lambda_{\max} (\mathbf{X}_{\mathsf{S}}^T \mathbf{X}_{\mathsf{S}}).
\end{equation*} 
\end{proposition}
\paragraph{Log-conditional with non-zero regularization term.} 
We now study log-conditional functions as in \eqref{eq:glm1AB}, under the assumption that the corresponding objective $l$ has a regularization term with parameter $\eta > 0$. We introduce additional notation to this end. For any feature set $\mathsf{T} \subseteq [n]$, denote with $\mathcal{P}_{\mathsf{T}}$ an operator that takes as input vectors $\mathbf{x} \in \mathbb{R}^n$, and it replaces all indices in $[n]\setminus \mathsf{T}$ of $\mathbf{x}$ by $0$. For any vector $\mathbf{x} \in \mathbb{R}^p$, we define $\mathbf{x}_{\mathsf{T}} \coloneqq \mathcal{P}_{\mathsf{T}}(\mathbf{x})$. We consider the following assumption on the distribution of the features.
\begin{assumption}[Assumption by \citet{DBLP:journals/jmlr/BahmaniRB13}]
\label{assumption_regularization}
For fixed constants $r,R>0$, we make the following assumption on the feature matrix $\mathbf{X} \in \mathbb{R}^{n \times p}$. The rows $\mathbf{x}$ of $\mathbf{X}$ are generated $i.i.d.$, such that the following additional conditions hold. For any set $\mathsf{T} \subseteq [p]$ of size $\absl{\mathsf{T}} \leq r$,
\begin{itemize}
    \item $\pnorm{\mathbf{x}_{\mathsf{T}}}{2} \leq R$;
    \item none of the matrices $\mathcal{P}_{\mathsf{T}}\expect{}{\mathbf{x} \mathbf{x}^{T}} \mathcal{P}_{\mathsf{T}}$ are the zero matrix.
\end{itemize}
\end{assumption}
Following this notation, define
\begin{equation*}
 \phi_{\max} \coloneqq \max_{\absl{\mathsf{T}} \leq k} \lambda_{\max}(\mathcal{P}_{\mathsf{T}}\mathbf{C} \mathcal{P}_{\mathsf{T}}) \qquad \phi_{\min} \coloneqq \min_{\absl{\mathsf{T}} \leq k}\lambda_{\max}(\mathcal{P}_{\mathsf{T}}\mathbf{C} \mathcal{P}_{\mathsf{T}}),
\end{equation*}
with $\lambda_{\max} (\cdot)$ the largest eigenvalue. The following corollary holds.
\begin{proposition}[Corollary 4 by \citet{DBLP:journals/corr/ElenbergKDN16}] Consider a function $l$ as in \eqref{eq:glm1A}, and suppose that $\eta > 0$. Suppose that Assumption \ref{assumption_regularization} holds with parameters $r,R$, and suppose that the number of samples $s$ is lower-bounded as
\begin{equation*}
s > \frac{R(\log r + r(1 + \log \frac{n}{k} - \log \delta))}{\phi_{\min} (1 + \varepsilon)\log (1 + \varepsilon) - \varepsilon}.
\end{equation*}
Then, with probability at least $1 - \delta $ the function $l$ is $(m,M)$-(smooth, restricted concave), for all $\beta$ with at most $r$ non-zero coefficients. The parameters $m$ and $M$ are defined as $m = q(1 + \varepsilon) \phi_{\max} + \eta$ and $M = \eta$, with $q$ a constant fulfilling $q \geq \max_i h^{-1}(\tau) Z'' (\beta, \mathbf{x}_i)$ for $h^{-1}(\cdot), Z(\cdot, \cdot)$ as in \eqref{eq:glm1AB}.
\end{proposition}
\paragraph{Log-conditional with no assumptions on the regularization term.} We now study log-conditional functions as in \eqref{eq:glm1AB}, with no additional assumption on the regularization term. For simplicity, we consider functions $l$ as in \eqref{eq:glm1AB} with $\eta = 0$. However, these results can easily be extended to the general case. The following lemma holds.
\begin{lemma}[Corollary 2 by \citet{DBLP:journals/corr/ElenbergKDN16}]
Denote with $r$ an upper-bound on the sparsity of the feature sets. Following the notation introduced above, suppose that the feature matrix $\mathbf{X}$ consists of samples drawn from a sub-Gaussian distribution  with parameter $\sigma^2$ and covariance matrix $\Sigma$. Then, for $\eta = 0$ the function $l$ as in \eqref{eq:glm1A} is $(M,m)$-(smooth, restricted concave) with parameters
\begin{equation*}
M =  \alpha_u \lambda_{\max}(\Sigma) \left ( 3 + \frac{2 n r}{s}  \right )  \quad \mbox{and} \quad m = \alpha_\ell -  \frac{c^2\sigma^2}{\alpha_\ell } \frac{k \log n}{s},
\end{equation*}
with high probability, for $s > 0$ sufficiently large. The constant $\alpha_\ell$ depends on $(\sigma^2, \Sigma)$ and $k$; the constant $\alpha_u$ yields $\alpha_u \geq \max_i h^{-1}(\tau)^{-1} Z'' (\boldsymbol \beta, \mathbf{x}_i)$, with $h^{-1}(\cdot), Z(\cdot, \cdot)$ as in \eqref{eq:glm1AB}.
\end{lemma}
\subsection{Restricted Strong Concavity and Smoothness of the \texorpdfstring{$R^2$}{} Objective} 
\label{appendix:proof_of_cor}
In this section, we prove guarantees for the $R^2$ objective. To this end, given a feature matrix consisting of $n$ features and $k$ observations, the $R^2$ objective with regularization can be written as
\begin{equation}
\label{objective:2}
    l(\boldsymbol \beta) = R^2_{\mathbf{X}, \mathbf{y}}(\boldsymbol \beta)= 1 - \frac{1}{k}\pnorm{(\mathbf{y} - \dprod{\mathbf{X}}{\boldsymbol \beta}}{2}^2 . 
\end{equation}
The following lemma, similar to Lemma \ref{cor:approx2}, holds.
\begin{proposition}
\label{cor:approx1}
The $R^2$ objective $l$ for the linear regression as in \eqref{objective:2}, is $(m,M)$-(restricted smooth, restricted strong concave) with parameters
\begin{equation*}
    m \coloneqq \min_{\mathsf{S}} \lambda_{\min} (\mathbf{X}_{\mathsf{S}}^T \mathbf{X}_{\mathsf{S}}) \ \mbox{and} \ M \coloneqq \max_{\mathsf{S}} \lambda_{\max} (\mathbf{X}_{\mathsf{S}}^T \mathbf{X}_{\mathsf{S}}).
\end{equation*}
\end{proposition}
This lemma can easily be extended to the case of an $R^2$ objective with regularization term.
\section{Adaptivity and the \textsc{pram} Model}
\label{appendix:PRAM}
Recall that the adaptivity is defined as follows~\citep{DBLP:conf/stoc/BalkanskiS18}.
\begin{definition}[Adaptivity]
\label{def:adaptivity}
Given an oracle $f$, an algorithm is $r$-adaptive if every query $q$ to the oracle $f$ occurs at a round $i \in [r]$ such that $q$ is independent of the answers $f(q')$ to all other queries $q'$ at round $i$.
\end{definition}
The notion of adaptivity is closely related to other models such as Parallel Random Access Machines (\textsc{pram}). The \textsc{pram} model consists of a set of processors, that communicate via a single shared memory and a memory access unit. The adaptivity extend to \textsc{pram} via the notion of depth. The depth is the number of parallel steps in an algorithm or the longest chain of dependencies. We remark that the \textsc{pram} model assumes that the input is loaded in memory, whereas the adaptive complexity model only assumes access to an oracle function.
\section{The \texorpdfstring{$\shuff{}$}{} Sub-Routine}
\label{appendix:randSeq}
\begin{algorithm}[t]
	\caption{$\shuffle{\mathsf{X}, \mathsf{S}}$}
	\label{alg:rndseq}
	    $\mathsf{A} \gets \emptyset$\; 
 	    \While{$\mathsf{X}  \neq \emptyset $\label{alg:beginWhileAlg3}}{
 	     Let $\{x_1, \ldots, x_n\}$ be a permutation of $\mathsf{X}$ chosen uniformly at random\;
 	     $j^* \gets \max \{ j \colon \mathsf{S} \cup \mathsf{A} \cup \{x_1,\ldots,x_i\}_{i \leq j} \in \ind{} \} $\;
 	     $\mathsf{A} \gets \mathsf{A} \cup \{x_1, \dots,x_{j^*}\}$\label{alg:sequence:alg}\;
 	    $\mathsf{X} \gets \{e \in X\setminus (\mathsf{S} \cup \mathsf{A}) \colon \mathsf{S} \cup \mathsf{A} \cup e  \in \ind{} \}$\label{alg:rndseq:Y1}\;
        \label{alg:endWhileAlg3}
        }
   	\textbf{return} $\mathsf{A}$\;
\end{algorithm}
The $\shuff{}$ sub-routine is presented in Algorithm \ref{alg:rndseq}. This algorithm correspond to Algorithm A by \citet{DBLP:journals/jcss/KarpUW88}. This algorithm solves the problem of sampling a maximum independent set of $\ind{}$ uniformly at random.
To our knowledge, no other algorithm is known for this problem, with better adaptivity than Algorithm \ref{alg:rndseq}. Given as input a ground set $\mathsf{X}$, a current solution $\mathsf{S}$, and a $p$-system $\ind{}$, this algorithm finds a random set $\mathsf{A}$ such that $\mathsf{S} \cup \mathsf{A}$ is a maximum independent set for $\ind{}$. This algorithm iteratively shuffles the set $\mathsf{X}$, and then it identifies the longest prefix of this sequence that can be added to $\mathsf{S}$, without violating side constraints. This prefix is then added to $\mathsf{A}$. This algorithm terminates when $\mathsf{S} \cup \mathsf{A} \in \ind{}$ is a maximal independent set.

This algorithm uses parallel calls to the independence oracle of $\ind{}$. In fact, the evaluations for the feasibility of prefixes of $\mathsf{A}$ can be preformed in parallel. Hence, with this algorithm the adaptivity of the independence oracle corresponds to the number of iterations until convergence. It is well-known that this algorithm converges after expected $\bigo{\sqrt{r}}$ iterations, with $r$ the rank of $\ind{}$ (see Theorem 6 by \citet{DBLP:journals/jcss/KarpUW88}). This implies that the adaptivity of the independence oracle is $\bigo{\sqrt{r}}$. Although it is not known if this upper-bound on the adaptivity is tight, it is known that there is no algorithm that finds a maximum independent set of $\ind{}$ with less then $\tilde{\Omega} (n^{1/3})$ rounds (see Theorem 7 by \citet{DBLP:journals/jcss/KarpUW88}).
\section{Parameter Tuning for Algorithm \ref{alg}}
\label{appendix:parameter_tuning}
If $l $ is the log-likelihood of a linear model, or the $R^2$ objective of a linear model with normalized response variable, then $m$ and $M$ can be estimated in terms of the design of the feature matrix (see Appendix \ref{appendix:B}-\ref{appendix:proof_of_cor}). These estimates need not be tight, and certifying bounds for $m$ and $M$ is $\mathsf{NP}$-hard \citep{DBLP:journals/tit/BandeiraDMS13}. In general, the constant $m/M$ in Line 4 of Algorithm \ref{alg} requires tuning by making multiple runs of the algorithm. We remark that other known parallel algorithms for feature selection also require estimates of these parameters \citep{DBLP:conf/nips/QianS19}.

Parallel algorithms that estimate the rank are known for several $p$-systems. For instance,  the rank of a graphic matroid can be estimated with parallel algorithms that compute spanning trees. Furthermore, parallel rank oracles are known for matroids that can be represented as independent sets of vectors in a given field \citep{DBLP:journals/iandc/BorodinGH82,DBLP:conf/fct/Chistov85,DBLP:journals/ipl/IbarraMR80,DBLP:journals/combinatorica/Mulmuley87}. These algorithms can also be used to estimate the rank of more complex constraints, such as the intersection of matroids or $p$-matchoids.
\section{Proof of Theorem \ref{thm:approx}}
\label{appendix:A}
We prove the following theorem.
\approx*
The proof of this theorem is based on a few lemmas and propositions, which we discuss in Appendix \ref{appendix:preliminaries} before proving Theorem \ref{thm:approx}. On a high level, the proof of Theorem \ref{thm:approx} is split into two separate cases. First we prove that Theorem \ref{thm:approx} holds when Algorithm \ref{alg} terminates after $\varepsilon^{-1}$ iterations of the outer While-loop of Algorithm \ref{alg}. Then, we prove Theorem \ref{thm:approx} under the assumption that Algorithm \ref{alg} finds a solution of size $k$. The first part of the proof is discussed in Appendix \ref{appendix:approxA} (see Theorem \ref{thm:approxA}), and the second case is discussed in Appendix \ref{appendix:approxB} (see Theorem \ref{thm:approxB}).
\subsection{Preliminary Results}
\label{appendix:preliminaries}
Our analysis is based on a few preliminary result, which we discuss in this section.
\begin{theorem}
\label{thm:singer++}
Suppose the $l$ is ($M$,$m$)-(smooth, strongly concave) on $\Omega_{2k}$. Then, for each subsets $\mathsf{S}, \mathsf{T} \subseteq [p]$ of size at most $k$ it holds
\begin{equation*}
   2M\sum_{j \in \mathsf{T}}f_{\mathsf{S}}(j) \geq \pnorm{\nabla l (\boldsymbol \beta ^{(\mathsf{S})})_{\mathsf{T}}}{2}^2 \geq 2m\sum_{j \in \mathsf{T}}f_{\mathsf{S}}(j).
\end{equation*}
\end{theorem}
\begin{proof}
We start proving the first inequality. Fix a point $j \in \mathsf{T}$. Then, for any scalar $\alpha$ it holds
\begin{align}
f_{\mathsf{S}}(j) = l(\boldsymbol\beta^{(\mathsf{S}\cup \{j\})}) - l(\boldsymbol\beta^{(\mathsf{S})}) & \geq l (\boldsymbol\beta^{(\mathsf{S})} + \alpha \mathbf{e}_{j}) - l(\boldsymbol\beta^{(\mathsf{S})}) & (\mbox{maximality of $\boldsymbol\beta^{(\mathsf{S}\cup \{j\})}$})\nonumber \\
& \geq \dprod{ \nabla l(\boldsymbol \beta^{(\mathsf{S})})}{ \alpha \mathbf{e}_{j} } -  \frac{M}{2} \alpha^2, & (\mbox{restricted smoothness})\label{eq:asdkjh}
\end{align}
By substituting $\alpha = \frac{1}{M}\dprod{ \nabla \oracle {\boldsymbol \beta^{(\mathsf{S})}}}{ \mathbf{e}_{j} }$ in \eqref{eq:asdkjh}, we get
\begin{equation*}
2M\sum_{j \in \mathsf{T}}f_{\mathsf{S}}(j) \geq  \sum_{j \in \mathsf{T}} \left \langle \nabla l(\boldsymbol \beta^{(\mathsf{S})}), \mathbf{e}_{j} \right \rangle^2 = \pnorm{\nabla l (\boldsymbol \beta ^{(\mathsf{S})})_{\mathsf{T}}}{2}^2,
\end{equation*}
and the first inequality follows. To conclude the proof, note that it holds
\begin{align}
& f_{\mathsf{S}}(j) = l(\boldsymbol \beta^{(\mathsf{S}\cup \{j\})}) - l (\boldsymbol \beta ^{(\mathsf{S})}) & \nonumber \\
& \leq \dprod{\nabla l (\boldsymbol \beta ^{(\mathsf{S})})}{\boldsymbol \beta^{(\mathsf{S}\cup \{j\})} - \boldsymbol \beta ^{(\mathsf{S})}} - \frac{m}{2} \pnorm{\boldsymbol \beta ^{(\mathsf{S}\cup \{j\})} - \boldsymbol \beta ^{(\mathsf{S})}}{2}^2 & (\mbox{restricted strong concavity}) \nonumber\\
    & \leq \max_{\mathbf{v} \colon \mathbf{v}_{(\mathsf{S} \cup \{j\}) = 0}}\dprod{\nabla l (\boldsymbol \beta ^{(\mathsf{S})})}{\mathbf{v} - \boldsymbol \beta ^{(\mathsf{S})}} - \frac{m}{2} \pnorm{\mathbf{v} - \boldsymbol \beta ^{(\mathsf{S})}}{2}^2 . & (\mbox{maximality of $\mathbf{v}$})\label{eq:new2}
\end{align}
By setting $\mathbf{v} = \boldsymbol \beta ^{(\mathsf{S})} + \frac{1}{m}\dprod{\nabla l(\boldsymbol \beta^{(\mathsf{S})})}{\mathbf{e}_j} $ in \eqref{eq:new2} we get
\begin{equation*}
\frac{\pnorm{\nabla l(\boldsymbol \beta ^{(\mathsf{S})})_{\mathsf{S}^*}}{2}^2}{2m} \geq l(\boldsymbol \beta^{(\mathsf{S}\cup \{j\})}) - l (\boldsymbol \beta ^{(\mathsf{S})}) = f_{\mathsf{S}} (j).
\end{equation*}
By taking the sum over all $j \in \mathsf{T}$ and rearranging we get
\begin{equation*}
\pnorm{\nabla l (\boldsymbol \beta ^{(\mathsf{S})})_{\mathsf{T}}}{2}^2 = \sum_{j \in \mathsf{T}}\frac{\pnorm{\nabla l(\boldsymbol \beta ^{(\mathsf{S})})_{\mathsf{S}^*}}{2}^2}{2m} \geq  \sum_{j \in \mathsf{T}}f_{\mathsf{S}} (j),
\end{equation*}
and the claim follows.
\end{proof}
In our analysis, we also use the following well-known result.
\begin{theorem}[Theorem 1 by \citet{DBLP:journals/corr/ElenbergKDN16}]
\label{thm:rajiv+}
Suppose the $l$ is ($M$,$m$)-(smooth, strongly concave) on $\Omega_{2k}$. Then, for each subsets $\mathsf{S}, \mathsf{T} \subseteq [p]$ of size at most $k$ it holds
\begin{equation*}
   2M f_{\mathsf{S}}(\mathsf{T}) \geq \pnorm{\nabla l (\boldsymbol \beta ^{(\mathsf{S})})_{\mathsf{T}}}{2}^2 \geq 2m f_{\mathsf{S}}(\mathsf{T}).
\end{equation*}
\end{theorem}
By combining Theorem \ref{thm:singer++} with Theorem \ref{thm:rajiv+}, we get the following corollary. 
\begin{corollary}
\label{cor:rajiv}
Suppose the $l$ is ($M$,$m$)-(smooth, strongly concave) on $\Omega_{2k}$. Then, for each subsets $\mathsf{S}, \mathsf{T} \subseteq [p]$ of size at most $k$ it holds
\begin{equation*}
   \frac{M}{m} f_{\mathsf{S}}(\mathsf{T}) \geq \sum_{j \in \mathsf{T}}f_{\mathsf{S}}(j) \geq \frac{m}{M} f_{\mathsf{S}}(\mathsf{T}).
\end{equation*}
\end{corollary}
We also make use of the following technical proposition.
\begin{proposition}[Proposition 2.2. by \citet{DBLP:journals/mp/NemhauserWF78}]
\label{prop:Fisher}
Let $\{x_1,\dots, x_m\}$ and $\{y_1,\dots, y_m\}$ be two sequences of non-negative real numbers. Suppose that it holds $\sum_j x_j \leq 1$ for all $j \in [m]$. Then,
\begin{equation*}
   \sum_{j = 1}^m y_j \geq  \sum_{j = 1}^m  x_j y_j.
\end{equation*}
\end{proposition}
\subsection{If Algorithm \ref{alg} Outputs a Maximum Independent Set}
\label{appendix:approxB}
We now prove Theorem \ref{thm:approx}, assuming that Algorithm \ref{alg} outputs a solution $\mathsf{S}$ of maximum size, before performing $\varepsilon^{-1}$ iterations of the outer While-loop of Algorithm \ref{alg}. Formally, we prove the following theorem.
\begin{theorem}
\label{thm:approxB}
Define the function $f$ as in \eqref{eq:def_f}, with a log-likelihood function that is $(M, m)$-(smooth, strongly concave) on $\Omega_{2r}$. Suppose that Algorithm \ref{alg} outputs a solution $\mathsf{S} ^*$ such that $\indspan{\mathsf{S}^*}= \emptyset $. Then,
\begin{equation*}
\frac{\expect{}{\falgo}}{\fopt} \geq \frac{1}{1 + p}\left (1 - \exp \left \{-(1 - \varepsilon)^2 \frac{m^3}{M^3}\right \} \right ) ,
\end{equation*}
for all $0 < \varepsilon < 1 $. Furthermore, in the specific case when $\ind{}$ is $r$-sparsity constraint over $[n]$, then,
\begin{equation*}
\frac{\expect{}{\falgo}}{\fopt} \geq 1 - \exp \left \{-(1 - \varepsilon)^2 \frac{m^2}{M^2}\right \} .
\end{equation*}
\end{theorem}
The proof of Theorem \ref{thm:approxB} is based on the following two additional lemma.
\begin{lemma}
\label{lemma:t}
At any point during the optimization process it holds
 \begin{equation*}
     (1 - \varepsilon )^{-1}\frac{r M}{m}t \geq 2m \left ( \fopt - f( \mathsf{S}) \right ),
 \end{equation*}
 with $\mathsf{S}$ the current solution.
\end{lemma}
\begin{proof}
Denote with $\bar{\mathsf{S}}$ a solution of size at most $\absl{\bar{\mathsf{S}}} \leq r$ maximizing $f(\mathsf{S} \cup \bar{\mathsf{S}})$, and let $\mathsf{T} \subseteq [n]$ be a set maximizing $\pnorm{\nabla l (\boldsymbol \beta ^{(\mathsf{S})})_{\mathsf{T}}}{2}^2$, such that $ \absl{\mathsf{T}}\leq r $ and $\mathsf{S}\cup \{s\}\in \ind{} $ for all $s \in \mathsf{T}$. Note that it holds
\begin{equation}
\label{eq:new_eqlemma}
    (1 - \varepsilon )^{-1}\frac{r M}{m}t = \frac{r}{\absl{\mathsf{T}}}\pnorm{\nabla l (\boldsymbol \beta ^{(\mathsf{S})})_{\mathsf{T}}}{2}^2  \geq \pnorm{\nabla l (\boldsymbol \beta ^{(\mathsf{S})})_{\mathsf{T}}}{2}^2,
\end{equation}
where the first inequality follows by the definition of $t$, and the second one follows since $\absl{\mathsf{T}} \leq r$. We first prove the claim when $t$ is updated at the beginning of each iteration of the outer While-loop of Algorithm \ref{alg}. It holds
\begin{align}
   & l(\boldsymbol  \beta^{(\mathsf{S}\cup \bar{\mathsf{S}})})  - l (\boldsymbol \beta ^{(\mathsf{S})}) & \nonumber \\
   & \leq \dprod{\nabla l (\boldsymbol \beta ^{(\mathsf{S})})}{\boldsymbol \beta^{(\mathsf{S}\cup \bar{\mathsf{S}})} - \boldsymbol \beta ^{(\mathsf{S})}} - \frac{m}{2} \pnorm{\boldsymbol \beta ^{(\mathsf{S}\cup \bar{\mathsf{S}})} - \boldsymbol \beta ^{(\mathsf{S})}}{2}^2 & (\mbox{restricted strong concavity})\nonumber \\
    & \leq \max_{\mathbf{v} \colon \mathbf{v}_{(\mathsf{S} \cup \bar{\mathsf{S}}) \neq 0}}\dprod{\nabla l (\boldsymbol \beta ^{(\mathsf{S})})}{\mathbf{v} - \boldsymbol \beta ^{(\mathsf{S})}} - \frac{m}{2} \pnorm{\mathbf{v} - \boldsymbol \beta ^{(\mathsf{S})}}{2}^2 . & (\mbox{maximality of $\mathbf{v}$}) \label{bold_v}
\end{align}
By setting $\mathbf{v} = \frac{1}{m}\boldsymbol \beta ^{(\mathsf{S})} + \nabla l(\boldsymbol \beta^{(\mathsf{S})})_{\bar{\mathsf{S}}} $ in the inequality above we get
\begin{align}
f(\fopt)- f( \mathsf{S}) & = l(\boldsymbol \beta^{(\mathsf{S}\cup \bar{\mathsf{S}})}) - l (\boldsymbol \beta ^{(\mathsf{S})}) & (\mbox{maximality of $\bar{\mathsf{S}}$ and monotonicity})\nonumber \\
& \leq \frac{\pnorm{\nabla l(\boldsymbol \beta ^{(\mathsf{S})})_{\bar{\mathsf{S}}}}{2}^2}{2m} & (\mbox{substituting $\mathbf{v}$ in \eqref{bold_v}})\nonumber \\
& \leq \frac{\pnorm{\nabla l (\boldsymbol \beta ^{(\mathsf{S})})_{\mathsf{T}}}{2}^2}{2m}. & (\mbox{maximality of $\mathsf{T}$}) \label{neq:eq_abcde}
\end{align}
The claim follows by combining \eqref{neq:eq_abcde} and \eqref{eq:new_eqlemma}. Suppose now that the current solution $\mathsf{S}$ is updated to $\mathsf{S}'$ during the inner While-loop of Algorithm \ref{alg}. Then, $f(\mathsf{S}) \geq f(\mathsf{S}')$ due to monotonicity, and the claim holds.
\end{proof}
\begin{lemma}
\label{lemma:marginal_contribution}
At any given time step, suppose that the current solution $\mathsf{S}$ is updated to $\mathsf{S} \cup \{a_1, \dots, a_{j^*} \}$, and define $\mathsf{S}_j = \mathsf{S} \cup \{a_1, \dots, a_j\}$ for all $j \in [j^*]$. Then it holds
\begin{equation*}
    \expect{}{f_{\mathsf{S}_{j- 1}} (\mathsf{S}_{j})} \geq (1 - \varepsilon)^2 \frac{m^2}{r M^2} \left ( \fopt - \expect{}{f( \mathsf{S}_{j- 1})} \right ) .
\end{equation*}
\end{lemma}
\begin{proof} Fix all random decisions of Algorithm \ref{alg} until the point $a_{j}$ is added to the current solution. Define the sets $\mathsf{X}_j^{\ind{}}\coloneqq \{e \in \mathsf{X} \setminus \{a_1, \dots, a_{j - 1} \} \colon \{a_1, \dots, a_{j - 1} \} \cup \{a \} \in \ind{} \}$.
Then, it holds
\begin{align}
\expect{a_{j}}{f_{\mathsf{S}_{j- 1}}(a_{j})} & \geq \frac{1}{2M} \expect{a_{j}}{\dprod{ \nabla l(\boldsymbol \beta^{(\mathsf{S}_{j- 1})})}{ \mathbf{e}_{a_{j}}}^2} & (\mbox{Theorem \ref{thm:singer++}})\nonumber \\ 
& \geq \frac{1}{2M} \pr{a_{j}}{\dprod{ \nabla l(\boldsymbol \beta^{(\mathsf{S}_{j- 1})})}{ \mathbf{e}_{a_{j}} }^2 \geq t} t & (\mbox{Markov's inequality}) \label{eq:new_eq_ex}
\end{align}
By design of the \shuff{} subroutine, each point $a_j$ is sampled uniformly at random from the set $\mathsf{X}_j^{\ind{}}$, i.e. $a_j\sim \mathcal{U}(\mathsf{X}_j^{\ind{}})$. Hence, it holds $\pr{a_{j}}{\dprod{ \nabla l(\boldsymbol \beta^{(\mathsf{S}_{j- 1})})}{ \mathbf{e}_{a_{j}} }^2 \geq t} = \absl{\mathsf{X}_{j - 1}} / \absl{\mathsf{X}_j^{\ind{}}}$. Combining this observation with \eqref{eq:new_eq_ex} we get
\begin{align*}
\expect{a_{j}}{f_{\mathsf{S}_{j- 1}}(a_{j})} 
& \geq \frac{1}{2M} \frac{\lvert \mathsf{X}_{j - 1} \rvert}{\absl{\mathsf{X}_j^{\ind{}}}}t, & (\mbox{by the sub-sampling procedure}) \nonumber \\
& \geq \frac{1}{2M} \frac{\lvert \mathsf{X}_{j - 1} \rvert}{\absl{\mathsf{X}}}t, & (\mbox{since $\mathsf{X}_j^{\ind{}} \subseteq \mathsf{X}$}) \\
& \geq (1 - \varepsilon) \frac{1}{2M} t & (\mbox{since $\absl{\mathsf{X}_{j-1}} \geq (1 - \varepsilon ) \absl{\mathsf{X}}$})\\
& \geq (1 - \varepsilon)^2 \frac{m^2}{r M^2} \left ( \fopt - f(\mathsf{S}_{j-1} ) \right ), & (\mbox{by Lemma \ref{lemma:t}})
\end{align*}
The claim follows by taking the expectation on both sides.
\end{proof}
Using this lemma, we can now prove Theorem \ref{thm:approxB}.
\begin{proof}[Proof of Theorem \ref{thm:approxB}]
Denote with $\{a_1, \dots, a_j\}$ the first $j$ points added to the solution $\mathsf{S}^*$, sorted in the order that they were added to it, and define the constant $c \coloneqq (1 - \varepsilon)^2 m^2/r M^2$. Using an induction argument on $j$, we prove that it holds
\begin{equation}
\label{eq:new_induction}
\frac{\expect{}{f(\{a_1, \dots,a_j\})}}{\fopt} \geq \left (1 - (1 - c)^j  \right ).
\end{equation}
The base case with $j = 0$ holds, due to the non-negativity of the function $f$. For the inductive case, we have that it holds
\begin{align*}
    \expect{}{f(\{a_1, \dots,a_j\})} & \geq \expect{}{f(\{a_1, \dots,a_{j-1}\})} + c \left ( \fopt - f(\{a_1, \dots,a_{j-1}\}) \right ) & (\mbox{Lemma \ref{lemma:marginal_contribution}})\\
    & \geq (1 - c)\left (1 - (1 - c)^{j-1}  \right ) \fopt + c \fopt  & (\mbox{by induction})\\
    & \geq \left (1 - (1 - c)^{j}  \right ) \fopt,
\end{align*}
and \eqref{eq:new_induction} holds. It follows that for $j = \absl{\mathsf{S}^*}$ we have
\begin{equation*}
    \frac{\expect{}{f(\mathsf{S}^*)}}{\fopt} \geq 1 - (1 - c)^{\absl{\mathsf{S}}} \geq 1 - \exp \left \{ - \frac{\absl{\mathsf{S}^*}}{k} c  \right \} .
\end{equation*}
In the case of a $r$-sparsity constraint we have that $\absl{\mathsf{S}^*} = r $, and the claim follows. In the case of a $p$-system constraint, we have that $\absl{\mathsf{S}^*}\geq pr$, hence
\begin{align*}
    \frac{\expect{}{f(\mathsf{S}^*)}}{\fopt} & \geq 1 - \exp \left \{ -(1 - \varepsilon)^2 p \frac{m^2}{M^2} \right \}  \geq \frac{1}{1 + p}\left (1 - \exp \left \{ -(1 - \varepsilon)^2 \frac{m^3}{M^3} \right \} \right ) ,
\end{align*}
and the claim also holds.
\end{proof}
\subsection{If Algorithm \texorpdfstring{\ref{alg}} Terminates after \texorpdfstring{$\varepsilon^{-1}$} Iterations}
\label{appendix:approxA}
We now prove Theorem \ref{thm:approx}, assuming that the \algo{} terminates after $\varepsilon^{-1}$ iterations of the outer While-loop of Algorithm \ref{alg}. Specifically, we prove the following theorem.
\begin{theorem}
\label{thm:approxA}
Define the function $f$ as in \eqref{eq:def_f}, with a log-likelihood function that is $(M, m)$-(smooth, strongly concave) on $\Omega_{2r}$. Suppose that Algorithm \ref{alg} terminates after $\varepsilon^{-1}$ iterations of the outer-While loop. Then,
\begin{equation*}
\frac{\expect{}{\falgo}}{\fopt} \geq \frac{1}{1 + p}\left (1 - \exp \left \{-(1 - \varepsilon)^2 \frac{m^3}{M^3}\right \} \right ) ,
\end{equation*}
for all $0 < \varepsilon < 1 $. Furthermore, in the specific case when $\ind{}$ is $r$-sparsity constraint over $[n]$, then,
\begin{equation*}
\frac{\expect{}{\falgo}}{\fopt} \geq 1 - \exp \left \{-(1 - \varepsilon)^2 \frac{m^2}{M^2}\right \} .
\end{equation*}
\end{theorem}
In order to prove this theorem, we introduce additional notation. We denote with $\mathsf{S}_i$ the current solution at the beginning of the $i$-th iteration of the outer While-loop of Algorithm \ref{alg}. Furthermore, denote with $\bar{\mathsf{S}} \subseteq \mathsf{T} \setminus \mathsf{S}_i$ a feasible set, such that $f(\bar{\mathsf{S}}\cup \mathsf{S}_i) = \fopt$, and denote with $a_j$ the $j$-th element added to the solution $\mathsf{S}$. The proof of this theorem is based on the following lemma.
\begin{lemma}
\label{lemma:approxA}
It holds 
\begin{equation*}
     \frac{M}{m}f_{\mathsf{S}_i}(\mathsf{S}_{i + 1}) + \sum_{e \in \bar{\mathsf{S}} \setminus \indspan{\mathsf{S}_i}}  \varepsilon f_{\mathsf{S}_{i}}(e) \geq \varepsilon \frac{m}{M} f_{\mathsf{S}_{i}}(\bar{\mathsf{S}} ).
\end{equation*}
\end{lemma}
\begin{proof}
Fix all random decision of Algorithm \ref{alg}, up to the $(i + 1)$-th iteration of the outer While-loop of Algorithm \ref{alg}. Let $\mathsf{T} \subseteq [n]$ be a set maximizing $ \pnorm{\nabla l (\boldsymbol \beta ^{(\mathsf{S}_i)})_{\mathsf{T}}}{2}^2$, such that $ \absl{\mathsf{T}} \leq r $  and $\mathsf{T} \subseteq \ind{}$. Due to the assumption on the stopping criterion, it holds $\mathsf{X} = \emptyset$ at the end of iteration $i$. This means that each point $j \in (\mathsf{T} \setminus \mathsf{S}_i) \cap \indspan{\mathsf{S}_i}$ was discarded at some point during the previous iteration. Denote with $\mathsf{U}_j$ the current solution when $j$ was discarded. Then, it holds 
\begin{align}
\label{eq:new_eq123}
(1 - \varepsilon) \frac{2 m}{r} \sum_{e \in \mathsf{T}}f_{\mathsf{S}_i}(e) \geq (1 - \varepsilon)\frac{m}{r M} \pnorm{\nabla l (\boldsymbol \beta ^{(\mathsf{S}_i)})_{\mathsf{T}}}{2}^2 = t 
\end{align}
where the first inequality follows by Theorem \ref{thm:singer++}, and the second one follows by the definition of $t$. Since the point $j$ was discarded and since $j \in \indspan{\mathsf{S}_i}$, then it must hold $t \geq \dprod{ \nabla l (\boldsymbol \beta_j^{(\mathsf{U}_{j})})}{ \mathbf{e}_j}^2$. Note also that by the RSC/RSM properties of the function $l$ it holds $\dprod{ \nabla l (\boldsymbol \beta^{(\mathsf{U}_{j})})}{ \mathbf{e}_j}^2 \geq \dprod{ \nabla l (\boldsymbol \beta^{(\mathsf{S}_{i+1})})}{ \mathbf{e}_j}^2$ \citep{DBLP:journals/corr/ElenbergKDN16}. Combining these observations with \eqref{eq:new_eq123} we get
\begin{align}
\label{eq:new_eq1}
(1 - \varepsilon) \frac{2 m}{r} \sum_{e \in \mathsf{T}}f_{\mathsf{S}_i}(e) &  \geq \dprod{ \nabla l (\boldsymbol \beta^{(\mathsf{S}_{i+1})})}{ \mathbf{e}_j}^2 \geq 2m f_{\mathsf{S}_{i + 1}}(j).
\end{align}
By taking the sum over all points $j\in (\mathsf{T} \setminus \mathsf{S}_i)\cap \indspan{\mathsf{S}_i}$ and rearranging, we get
\begin{align}
(1 - \varepsilon) \sum_{e \in \mathsf{T}}f_{\mathsf{S}_i}(e) & \geq  (1 - \varepsilon)\frac{\absl{(\mathsf{T} \setminus \mathsf{S}_i)\cap \indspan{\mathsf{S}_i}}}{r} \sum_{e \in \mathsf{T}}f_{\mathsf{S}_i}(e) & (\mbox{by definition of $r$})\nonumber \\
& \geq \sum_{j \in (\mathsf{T} \setminus \mathsf{S}_i)\cap \indspan{\mathsf{S}_i}}\bar{f}_{\mathsf{S_{i + 1}}}(j)& (\mbox{it follows from \eqref{eq:new_eq1}}) \label{eq:Appendix1}
\end{align}
By rearranging \eqref{eq:Appendix1} we get
\begin{align}
     \sum_{e \in \mathsf{S}_{i+1}} f_{\mathsf{S}_i}(e) & \geq  \sum_{e \in (\mathsf{T} \setminus \mathsf{S}_i)\cap \indspan{\mathsf{S}_i}}  \varepsilon f_{\mathsf{S}_{i}}(e)\nonumber \\
      & \geq \sum_{e \in \bar{\mathsf{S}} \cap \indspan{\mathsf{S}_i}}\varepsilon f_{\mathsf{S}_{i}}(e) & (\mbox{by the definition of $\overline{\mathsf{S}}$})\nonumber \\ 
      & \geq \sum_{e \in \bar{\mathsf{S}}}  \varepsilon f_{\mathsf{S}_{i}}(e) - \sum_{e \in \bar{\mathsf{S}} \setminus \indspan{\mathsf{S}_i}}  \varepsilon f_{\mathsf{S}_{i}}(e) & (\mbox{by linearity})\label{eq_final_new}
\end{align}
Hence, it holds
\begin{align*}
    \frac{M}{m}f_{\mathsf{S}_i}(\mathsf{S}_{i + 1}) & \geq \sum_{e \in \mathsf{S}_{i+1}} f_{\mathsf{S}_i}(e) & (\mbox{Corollary \ref{cor:rajiv}})\\
    & \geq \sum_{e \in \bar{\mathsf{S}}}  \varepsilon f_{\mathsf{S}_{i}}(e) - \sum_{e \in \bar{\mathsf{S}} \setminus \indspan{\mathsf{S}_i}}  \varepsilon f_{\mathsf{S}_{i}}(e) & (\mbox{it follows from \eqref{eq_final_new}})\\ 
    & \geq \varepsilon \frac{m}{M}  f_{\mathsf{S}_{i}}(\bar{\mathsf{S}}) - \sum_{e \in \bar{\mathsf{S}} \setminus \indspan{\mathsf{S}_i}}  \varepsilon f_{\mathsf{S}_{i}}(e). & (\mbox{Corollary \ref{cor:rajiv}})
\end{align*}
The claim follows by rearranging.
\end{proof}
In order to continue with the proof, we also use the following lemma.
\begin{lemma}
\label{lemma:approx_new_Dj}
It holds
\begin{equation*}
     pf(\mathsf{S}_{i}) \geq (1 - \varepsilon)\frac{m^2}{M^2} \sum_{e \in \bar{\mathsf{S}} \setminus \indspan{\mathsf{S}_i}}  f_{\mathsf{S}_{i}}(e).
\end{equation*}
\end{lemma}
\begin{proof}
Fix an index $j \leq \absl{\mathsf{S}_i}$, and fix all random decisions of Algorithm \ref{alg} until the point $a_{j} $ is added to the current solution. For each element $a_j$, define the set $\mathsf{D}_j \coloneqq (\bar{\mathsf{S}} \cap \indspan{\{a_1, \dots, a_{j - 1}\}}) \setminus (\bar{\mathsf{S}} \cap \indspan{\{a_1, \dots, a_{j }\}})$. Note that these sets consist of all points in $\bar{\mathsf{S}}$ that yield a feasible solution when added to $\{a_1, \dots, a_{j - 1}\}$, but that violate side constraints when added to $\{a_1, \dots, a_{j }\}$. Note also that it holds $\mathsf{D}_1 \cup \dots \cup \mathsf{D}_j = \bar{\mathsf{S}} \setminus \indspan{\{a_1, \dots, a_{j }\}}$. Furthermore, define the sets $$\mathsf{X}_{j}^{\ind{}}\coloneqq \{e \in \mathsf{X} \setminus \{a_1, \dots, a_{j-1} \} \colon \{a_1, \dots, a_{j-1} \} \cup \{a \} \in \ind{} \}.$$
Then, it holds
\begin{align}
\expect{a_{j}}{f_{\{a_1, \dots, a_{j-1} \}}(a_{j})}  & \geq \frac{1}{2M} \expect{a_{j}}{\dprod{ \nabla l(\boldsymbol \beta^{(\{a_1, \dots, a_{j-1} \})})}{ \mathbf{e}_{a_{j}}}^2} & (\mbox{by Lemma \ref{thm:singer++}})\nonumber \\
& \geq \frac{1}{2M} \pr{a_{j}}{\dprod{ \nabla l(\boldsymbol \beta^{(\{a_1, \dots, a_{j-1} \})})}{ \mathbf{e}_{a_{j}} }^2 \geq t}t  & (\mbox{by Markov})\label{eq:new_appendix}
\end{align}
Note that the \shuff{} subroutine samples points $a_j$ uniformly at random $a_j\sim \mathcal{U}(\mathsf{X}_j^{\ind{}})$. Hence, \begin{equation}
\label{eq:new_eq_proof_eps}
\pr{a_{j}}{\dprod{ \nabla l(\boldsymbol \beta^{(\mathsf{S}_{j- 1})})}{ \mathbf{e}_{a_{j}} }^2 \geq t} = \frac{\absl{\mathsf{X}_{j - 1}}}{ \absl{\mathsf{X}_j^{\ind{}}}} \geq \frac{\absl{\mathsf{X}_{j - 1}}}{ \absl{\mathsf{X}}} ,
\end{equation}
where the last inequality holds, since $  \mathsf{X}_j^{\ind{}}\subseteq \mathsf{X}$. Combining this observation with \eqref{eq:new_appendix} we get
\begin{align}
\expect{a_{j}}{f_{\{a_1, \dots, a_{j-1} \}}(a_{j})}  &  \geq \frac{1}{2M} \frac{\lvert \mathsf{X}_{j} \rvert}{\absl{\mathsf{X}}}t & (\mbox{it follows by \eqref{eq:new_eq_proof_eps}})\nonumber \\
&\ \  \geq \frac{1 - \varepsilon}{2M} t & (\mbox{$\absl{\mathsf{X}_{j-1}} \geq (1 - \varepsilon ) \absl{\mathsf{X}}$})\nonumber \\ 
&\ \  = \frac{1 - \varepsilon}{2M^2} \frac{m}{\absl{\mathsf{T}}} \pnorm{\nabla l (\boldsymbol \beta ^{(\{a_1, \dots, a_{j-1} \})})_{\mathsf{T}}}{2}^2 & (\mbox{by the definition of $t$})\nonumber 
\\ 
&\ \  \geq \frac{1 - \varepsilon}{2M^2} \frac{m}{\absl{\mathsf{D}_j}} \pnorm{\nabla l (\boldsymbol \beta ^{(\{a_1, \dots, a_{j-1} \})})_{\mathsf{D}_{j}}}{2}^2, & (\mbox{$\mathsf{T}$ is maximal})\nonumber
\end{align}
By taking the expected value on both sides in the chain of inequalities above, we get 
\begin{equation}
\label{eq:new_lemma123456}
\sum_{j \leq \absl{\mathsf{S}_{i}}}\absl{\mathsf{D}_j} \expect{a_{j}}{\bar{f}_{\{a_1, \dots, a_{j-1} \}}(a_{j})} \geq (1 - \varepsilon)\frac{m}{2M^2} \sum_{j \leq \absl{\mathsf{S}_{i}}} \expect{}{ \pnorm{\nabla l (\boldsymbol \beta ^{(\mathsf{S}_i)})_{\mathsf{D}_j}}{2}^2 }.
\end{equation}
In order to continue with the proof, we give an upper-bound on the size of the sum $\sum_j \absl{\mathsf{D}_j}$. To this end, note that the set $\mathsf{S}_i$ is a maximum independent set over the ground set
\begin{equation*}
  \mathsf{S}_i \cup \left ( \mathsf{D}_1 \cup \dots \cup \mathsf{D}_{\absl{\mathsf{S}_i}} \right ) = \mathsf{S}_i \cup \left ( \bar{\mathsf{S}} \setminus \indspan{\mathsf{S}_i} \right ).   
\end{equation*}
In fact, the set $\mathsf{S}_i$ is independent by definition, and that any point $s\in  \left ( \bar{\mathsf{S}} \setminus \indspan{\mathsf{S}_i} \right ) \setminus \mathsf{S}_i$ yields $\mathsf{S}_i \cup \{ s\} \notin \ind{}$. Hence $\mathsf{S}_i$ is a maximum independent set as claimed. Note also that $\mathsf{D}_1 \cup \dots \cup \mathsf{D}_{\absl{\mathsf{S}_i}}\subseteq \bar{\mathsf{S}}$ is an independent set, due to the subset-closure of $\ind{}$. Since $\ind{}$ is a $p$-system, then it holds
\begin{equation}
\label{eq1:claim2:thm1}
    \absl{\mathsf{D}_1} + \dots + \absl{\mathsf{D}_{\absl{\mathsf{S}_i}} } = \absl{\mathsf{D}_1 \cup \dots \cup \mathsf{D}_{\absl{\mathsf{S}_i}} } \leq p \absl{\mathsf{S}_i}.
\end{equation}
Hence, it holds
\begin{align}
p \sum_{j \leq \absl{\mathsf{S}_{i}}}  \expect{a_{j}}{f_{\{a_1, \dots, a_{j-1} \}}(a_{j})} & \geq \sum_{j \leq \absl{\mathsf{S}_{i}}}\absl{\mathsf{D}_j} \expect{a_{j}}{f_{\{a_1, \dots, a_{j-1} \}}(a_{j})}  & (\mbox{it follows by \eqref{eq1:claim2:thm1}}) \nonumber \\
& \geq (1 - \varepsilon)\frac{m}{2M^2} \sum_{j \leq \absl{\mathsf{S}_{i}}} \expect{}{ \pnorm{\nabla l (\boldsymbol \beta ^{(\mathsf{S}_i)})_{\mathsf{D}_j}}{2}^2 } & (\mbox{it follows by \eqref{eq:new_lemma123456}}) \nonumber \\
& \geq (1 - \varepsilon)\frac{m^2}{M^2} \sum_{e \in \mathsf{D}_1 \cup \dots \cup \mathsf{D}_{\absl{\mathsf{S}_i}}} f_{\mathsf{S}_i}(e) & (\mbox{by Theorem \ref{thm:singer++}}). \nonumber  
\end{align}
The claim follows since $\mathsf{D}_1 \cup \dots \cup \mathsf{D}_{\absl{\mathsf{S}_i}} = \bar{\mathsf{S}} \setminus \indspan{\mathsf{S}_i}$.
\end{proof}
We now have all necessary tools to prove Theorem \ref{thm:approxA}.
\begin{proof}[Proof of Theorem \ref{thm:approxA}]
We first prove the claim, assuming that $\ind{}$ is a general $p$-system. In this case, by combining Lemma \ref{lemma:approxA} with Lemma \ref{lemma:approx_new_Dj} it holds
\begin{equation}
\label{eq:Appendix2_new_abc}
     \frac{M}{m} \expect{}{f_{\mathsf{S}_i}(\mathsf{S}_{i + 1})} + \varepsilon p \frac{M^2}{m^2}\expect{}{f(\mathsf{S}_{i})} \geq \varepsilon (1 - \varepsilon)\frac{m}{M} \expect{}{f_{\mathsf{S}_{i}}(\bar{\mathsf{S}} )}.
\end{equation}
To continue, define the constant $c = (1 - \varepsilon)m^3/M^3$. We prove by induction on $i$ that it holds
\begin{equation}
\label{eq:Appendix2}
    ( 1 + \varepsilon i p)\expect{}{f(\mathsf{S}_{i})} \geq (1 - (1 - \varepsilon c )^i ) \fopt .
\end{equation}
The base case with $\mathsf{S}_0 = \emptyset $ trivially follows, since the function $f$ is non-negative. For the inductive case, suppose that the claim holds for $\expect{}{f(\mathsf{S}_{i -1})}$. Then,
\begin{align*}
    ( 1 + \varepsilon i p) &  \expect{}{f(\mathsf{S}_{i})} & \nonumber \\
    & \geq \expect{}{f(\mathsf{S}_{i})} + \varepsilon i p  \expect{}{f(\mathsf{S}_{i-1})} & (\mbox{by monotonicity})\nonumber \\
    & \geq \expect{}{f(\mathsf{S}_{i -1})} + \varepsilon c \expect{}{f_{\mathsf{S}_i}(\bar{\mathsf{S}})} + \varepsilon (i-1) p  \expect{}{f(\mathsf{S}_{i-1})} & (\mbox{it follows by \eqref{eq:Appendix2_new_abc}})\nonumber  \\
    & \geq \left ( 1 - \varepsilon c \right ) \expect{}{f(\mathsf{S}_{i -1})} + \varepsilon c \fopt  + \varepsilon (i-1) p  \expect{}{f(\mathsf{S}_{i-1})} & (\mbox{by monotonicity})\nonumber  \\
    & \geq ( 1 - \varepsilon c ) (1 - (1 - \varepsilon c )^{i-1} ) \fopt + \varepsilon c \fopt & (\mbox{by induction})\nonumber  \\
    & \geq (1 - (1 - \varepsilon c )^i ) \fopt .  &  
\end{align*}
Hence, \eqref{eq:Appendix2} holds. It follows that 
\begin{align*}
    \expect{}{\falgo}  & =  \expect{}{f(\mathsf{S}_{\left \lfloor 1/\varepsilon \right \rfloor})} & (\mbox{by the stopping criterion}) \\
    & \geq \frac{1}{ 1 + \varepsilon \left \lfloor 1/\varepsilon \right \rfloor p} \left (1 - \left (1 - \varepsilon (1 - \varepsilon) \frac{m^3}{M^3} \right )^{\left \lfloor 1/\varepsilon \right \rfloor} \right ) \fopt & (\mbox{it follows by \eqref{eq:Appendix2}}) \\ 
    & \geq \frac{1}{ 1 + p} \left (1 - \exp \left \{ - (1 - \varepsilon) \frac{m^3}{M^3} \right \} \right ) \fopt, & 
\end{align*}
as claimed.

We conclude by proving the claim in the special case that $\ind{}$ is a $r$-sparsity constraint. Since the algorithm terminates before a solution of size $r$ is found, we have that $\indspan{\mathsf{S}_i} = [n]\setminus \mathsf{S}_i$ for all iterations $i$. Hence, $\mathsf{D}_1 \cup \dots \cup \mathsf{D}_i = \emptyset$ and Lemma \ref{lemma:approxA} yields
\begin{equation*}
     \expect{}{f_{\mathsf{S}_i}(\mathsf{S}_{i + 1})} \geq \varepsilon \frac{m^2}{M^2} \expect{}{f_{\mathsf{S}_{i}}(\mathsf{S}^* )}.
\end{equation*}
With this inequality, we can use an inductive argument similar to the proof for the general case, and obtain an improved lower-bound on the solution quality.
\end{proof}
\section{Proof of Theorem~\ref{thm:run_time}}
\label{appendix:adptivity_run_time}
We conclude by giving upper-bounds on the run time and adaptivity for Algorithm \ref{alg}. Recall that the notion of adaptivity is given in Definition \ref{def:adaptivity}. The following theorem holds.
\runtime*
In order to prove this result, we use the following well-known estimate on the number of adaptive rounds of the \shuff{} sub-routine (see Appendix \ref{appendix:randSeq}).
\begin{theorem}[Theorem 6 by \citet{DBLP:journals/jcss/KarpUW88}]
\label{thm_Karp}
Algorithm \ref{appendix:randSeq} terminates after expected $\bigo{\sqrt{r}}$ steps, with $r$ the rank of the independent system $\ind{}$.
\end{theorem}
Note that this theorem implies that the number of adaptive rounds of the independence oracle for Algorithm \ref{appendix:randSeq} is $\bigo{\sqrt{r}}$ in expected value. In fact, in each step of Algorithm \ref{appendix:randSeq}, queries to the independence oracle can be performed in parallel. Using this result, we can now prove Theorem \ref{thm:run_time}.
\begin{proof}[Proof of Theorem \ref{thm:run_time}]
We first give upper-bounds for the oracle function that accesses $\nabla l(\cdot)$. To this end, observe that there are two While-loops in Algorithm \ref{alg}. The outer while-loop terminates after at most $\varepsilon^{-1}$ iteration. The inner While-loop terminates after
$\bigo{\varepsilon^{-1}\log n}$ iterations, since at each iteration the size of $\mathsf{X}$ decreases at least of a multiplicative factor of $1 - \varepsilon$. Hence, the rounds of calls to the oracle function is $\bigo{\varepsilon^{-2}\log n}$. Furthermore, at each iteration of the inner While-loop, at most $r$ parallel calls to $\nabla l(\cdot)$ are preformed. It follows that the total number of oracle calls is $\bigo{\varepsilon^{-2}r \log n}$.

We now estimate the number of adaptive rounds and run time for the calls to the independence oracle. To this end, note that is oracle is called by the $\shuff{}$ sub-routine,and it is also evaluated $nr$ times in parallel during the inner While-loop of Algorithm \ref{alg}. From Theorem \ref{thm_Karp} it follows that the number of adaptive rounds is $\bigo{\varepsilon^{-2}\sqrt{r}\log n}$, and that the total number of calls to the oracle function is $\bigo{\varepsilon^{-2}n r \log n}$ as claimed.
\end{proof}
\section{Feature selection on Non-volatile Memory (NVM)} \label{appendix:mcas}
The emergence of CPU-attached persistent memory technology, such as Intel’s Optane Non-Volatile Memory (NVM), has opened opportunities for running large datasets on a single server. A single server may have up to 6TB of NVM, that can be accessed through the memory bus. The use of NVM compared to DRAM is beneficial for two reasons. First, due to its large capacity, it is suitable to handle large datasets. Secondly, as NVM is non-volatile, it can easily support fault-tolerance and recovery of the computation when a server crashes. Furthermore, even though NVM latency is slightly slower than DRAM, it has much better latency than SSD of orders of magnitude.

For running our Python code on NVM, we used MCAS \citep{MCAS1,MCAS_ADO} (Memory Centric Active Storage), which is an advanced client-server in-memory object store designed from the ground up to leverage persistent memory. MCAS supports \say{pushdown} operations on the client-side which are termed Active Data Objects (ADO), and has a Python plugin that allows zero-copy for Numpy data access. Our benchmark for evaluating feature selection with persistent memory is done by integrating the algorithms to MCAS as an ADO using the above mentioned Python plugin. We use an Intel Xeon Gold  6248 server with 80 CPUs at 2.50GHz. The server is equipped with 384GB DDR4 DRAM and 1512GB Optane DC.

By experimenting with the same datasets with the \omp{} and \algo{}, we observed no significant variation in run time compared to our DRAM implementation. This is due to the fact that the time required to copy data from NVM to DRAM is negligible, in comparison with the compute time of training phases. However, when performing tasks on the DRAM that use more memory than the DRAM capacity, we might observe a significant decrease in the performance. For this reason, in future work we intend to investigate the trade-off between training time on different media (NVM versus DRAM and SSD) of tasks that use more memory than the DRAM capacity.
\section{The ProPublica COMPAS Dataset} \label{appendix:pro_publica}
The ProPublica COMPAS dataset was constructed in 2016, using data of defendants from Broward County, FL, who had been arrested in 2013 or 2014 and assessed with the COMPAS risk screening system. ProPublica then collected data on future arrests for these defendants through the end of March 2016, in order to study how the COMPAS score predicted recidivism \citep{doi:10.1177/0049124118782533123}. Based on its analysis, ProPublica concluded that the COMPAS risk score was racially biased \citep{doi:10.1177/0049124118782533}. 

The ProPublica COMPAS data has become one of the key bench-marking datasets for testing algorithmic fairness definitions and procedures \citep{DBLP:conf/kdd/Corbett-DaviesP17,chouldechova2017fair,corbett2017algorithmic,cowgill2019economics,rudin2018age,DBLP:conf/www/ZafarVGG17,DBLP:conf/nips/ZafarVGGW17}. However, \citet{DBLP:journals/corr/abs-2106-05498} notes that there are inaccuracies in the COMPAS dataset. For instance, COMPAS race categories lack Native Hawaiian or Other Pacific Islander, and it redefines Hispanic as race instead of ethnicity.

This dataset consists of the following features: \say{number of prior criminal offenses}, \say{arrest charge description}, \say{charge degree}, \say{number of juvenile felony offenses}, \say{juvenile misdemeanor offenses}, \say{other juvenile offenses}, \say{age}, \say{sex} and \say{race} of the defendant. The dataset also contains information on whether the defendant recidivated or not.

\end{document}